%% file: main.tex
\pdfoutput=1
\documentclass{article} %
\usepackage{iclr2027_conference,times}

\input{math_commands.tex}

\usepackage{enumitem}
\usepackage{hyperref}
\usepackage{url}
\usepackage{graphicx}
\usepackage{booktabs}
\usepackage{siunitx}
\usepackage{threeparttable}
\usepackage{subcaption}
\usepackage{placeins}
\usepackage{amssymb}
\definecolor{ablPure}{RGB}{204,120,188}
\definecolor{ablFree}{RGB}{222,143,5}
\definecolor{ablDual}{RGB}{1,115,178}

\newcommand{\modelname}{dIon}

\makeatletter
\newcommand{\DeclareDataset}[3]{\expandafter\gdef\csname dataset@#1@full\endcsname{#2}\expandafter\gdef\csname dataset@#1@short\endcsname{#3}}
\newcommand{\DatasetFull}[1]{\csname dataset@#1@full\endcsname}
\newcommand{\DatasetShort}[1]{\csname dataset@#1@short\endcsname}
\makeatother
\DeclareDataset{bacterialssl}{PXD010000 bacterial corpus}{PXD010000}
\DeclareDataset{bacterialbenchmark}{PXD010613 bacterial corpus}{PXD010613}
\DeclareDataset{ninespecies}{NineSpecies V2}{NineSpecies}
\DeclareDataset{kingdoms}{Kingdoms proteome landscape}{Kingdoms}
\DeclareDataset{mskbfinal}{MSKB-final de novo corpus}{MSKB-final}
\DeclareDataset{kingdomsdenovo}{Kingdoms run-disjoint de novo test}{Kingdoms de novo}
\DeclareDataset{kitchensinkvfive}{\modelname{}-de-novo-labeled-v1}{DNL-v1}
\DeclareDataset{sqa}{19-PXD human Orbitrap HCD corpus}{SQA-19PXD}
\DeclareDataset{hye}{PXD024584 HYE mixture corpus}{HYE}
\DeclareDataset{bacterialchim}{PXD010613-derived chimericity task}{Bacterial chim.}
\DeclareDataset{oxidizedmet}{PXD010613-derived oxidized-methionine task}{OxMet}
\DeclareDataset{massivekbvone}{MassIVE-KB v1}{MassIVE-KB v1}
\DeclareDataset{massivekbvtwo}{MassIVE-KB v2.0.15}{MassIVE-KB v2.0.15}
\DeclareDataset{prospect}{PROSPECT}{PROSPECT}
\input{code/generated/datasets/dataset_facts.tex}

\input{code/generated/tables/result_macros.tex}
\input{code/generated/tables/result_macros_ablation.tex}

\title{dIon\\ Fragmentation-Based Invariance for Self-Supervised Learning of Tandem Mass Spectra}

\author{Alfred Nilsson \\
Science for Life Laboratory, KTH Royal Institute of Technology \\
Stockholm, Sweden \\
\And
Joel Lapin \\
Computational Mass Spectrometry, TUM School of Life Sciences \\
Technical University of Munich, Freising, Germany \\
\And
Samuel H. Payne \\
Biology Department, Brigham Young University \\
Provo, Utah 84602, United States \\
\AND
Mathias Wilhelm \\
Computational Mass Spectrometry, TUM School of Life Sciences \\
Technical University of Munich, Freising, Germany \\
Munich Data Science Institute, Technical University of Munich \\
Garching, Germany \\
\And
Lukas Käll\thanks{Corresponding author: \texttt{lukas.kall@scilifelab.se}} \\
Science for Life Laboratory, KTH Royal Institute of Technology \\
Stockholm, Sweden
}

\iclrfinalcopy
\begin{document}
\raggedbottom

\maketitle
\lhead{Preprint}

\input{sections/abstract}

\input{sections/01_introduction}
\input{sections/02_related_work}
\input{sections/03_method}
\input{sections/04_experiments}
\input{sections/05_discussion}
\input{sections/06_conclusion}

\clearpage
\subsection*{AI use statement}

We used generative AI tools in this work and describe their role below. All
AI-assisted output was reviewed by the authors, and we take responsibility for
the final content of this paper, including its text, claims, figures and
released artifacts.

\textbf{Research ideation and experimental design.}
The contribution of this paper, the fragmentation-based invariance and the
dual latent-prediction objective that imposes it, was conceived by the
authors. We used large language models, through Anthropic's Claude Code and
OpenAI's Codex, to help design some of the auxiliary spectrum-level tasks reported in
Section~\ref{sec:results} and Appendix~\ref{app:downstream_tasks}. The authors
selected, specified and validated the final task definitions.

\textbf{Code.}
The model, training and evaluation code was written by the authors. We used
the same tools to refactor, document and debug that code, and to implement
dataset format conversions and preprocessing. All AI-assisted code was
reviewed and tested by the authors, and every reported number comes from a run
of the reviewed code.

\textbf{Writing.}
We used the same tools for drafting and revision, including reorganizing
sections, condensing text to the page limit and rewriting passages, and for
language correction alongside Grammarly. Every sentence of the submitted
manuscript was read and approved by the authors, and no text was included that
the authors had not checked against the underlying results.

\textbf{Related work.}
Most of the work we cite was known to us from our own research in these
fields. In a few cases, we used the same tools to find related work.

We did not use generative AI to produce, alter or select experimental results,
and no AI system is an author of this work.

\subsection*{Ethics statement}

Every corpus used here was collected, approved and publicly released by an
earlier study, and we accessed each one through its public repository under the
terms of that release. We collected no new samples and performed no experiments
on biological material. Some corpora derive from human samples; we use only the
released fragment spectra, we make no attempt to recover donor identity, and we
add no information that the original depositors did not publish. Accessions and
the originating publications are given in Appendix~\ref{app:datasets} so that
credit follows the data.

We see a wider benefit in this mode of work. Public repositories already hold
far more tandem mass spectra than database search can confidently identify, and
the unassigned majority is ordinarily discarded. Learning without peptide
labels extracts further value from measurements that have already been made and
paid for, rather than requiring new acquisition, and the resulting encoder is
reused across several downstream tasks. Against that we set the cost of
pretraining, which used 64 accelerators for approximately three days; we report
that cost in Appendix~\ref{app:architecture_training} and note that it is
incurred once and amortized over the transfer experiments reported here and any
later reuse of the released model.

De novo peptide sequencing is a general analytical capability, and we are not
aware of a misuse pathway specific to this work beyond those already present in
computational proteomics.

\subsection*{Reproducibility statement}

The pretraining objective and both latent-prediction tasks are specified in
Section~\ref{sec:method} and Appendix~\ref{app:ssl_objective}. Encoder
architecture, optimization, and the full training configuration are given in
Appendix~\ref{app:architecture_training}, with compute in
Appendix~\ref{app:compute}. Dataset manifests, split construction and the
filtering rules separating \DatasetShort{bacterialssl} from every evaluation
corpus are in Appendix~\ref{app:datasets}. The evaluation protocol, readouts
and decoding settings are in Appendices~\ref{app:embedding_analysis}
and~\ref{app:denovo_decoding}, and the handling of the external de novo models
in Appendix~\ref{app:kingdoms_eval}. Code is available at \url{https://github.com/statisticalbiotechnology/dIon}.
The pretrained \modelname{} encoder and the fine-tuned de novo checkpoints
are available at
\url{https://huggingface.co/alfred-n/dIon}.

\subsubsection*{Acknowledgments}
Computational resources were provided by the National Academic Infrastructure
for Supercomputing in Sweden (NAISS), funded by the Swedish Research Council,
on the Arrhenius system at NSC. Computational resources for early development
were also provided on the Berzelius system funded by the Knut and Alice
Wallenberg foundation and operated by NAISS.

\bibliography{references}
\bibliographystyle{iclr2027_conference}

\appendix
\input{appendix/A_datasets}
\input{appendix/B_architecture_training}
\input{appendix/C_embed_eval}

\input{appendix/D_downstream_transfer}
\input{appendix/E_denovo}

\input{appendix/F_training_diagnostics}

\end{document}

%% file: math_commands.tex
\usepackage{amsmath,amsfonts,bm}

\def\eqref#1{equation~\ref{#1}}

\def\1{\bm{1}}

\DeclareMathAlphabet{\mathsfit}{\encodingdefault}{\sfdefault}{m}{sl}
\SetMathAlphabet{\mathsfit}{bold}{\encodingdefault}{\sfdefault}{bx}{n}

\newcommand{\softmax}{\mathrm{softmax}}

%% file: code/generated/datasets/dataset_facts.tex
\newcommand{\DSBacterialTrain}{9,234,199}
\newcommand{\DSBacterialVal}{467,723}
\newcommand{\DSBacterialGeomVal}{51,770}
\newcommand{\DSNineSpeciesVal}{6,119}
\newcommand{\DSKingdomsVal}{658,738}
\newcommand{\DSBacterialTest}{518,804}
\newcommand{\DSNineSpeciesTest}{182,393}
\newcommand{\DSKingdomsTest}{4,112,590}
\newcommand{\DSBacterialTestChargeTwoToFour}{499,617}
\newcommand{\DSNineSpeciesTestChargeTwoToFour}{181,852}
\newcommand{\DSKingdomsTestChargeTwoToFour}{4,087,062}
\newcommand{\DSMSKBTrain}{1,989,996}
\newcommand{\DSMSKBVal}{10,004}
\newcommand{\DSMSKBTest}{200,000}
\newcommand{\DSKitchenSinkTrain}{5,275,467}
\newcommand{\DSKitchenSinkVal}{170,140}
\newcommand{\DSKitchenSinkTest}{345,414}
\newcommand{\DSSQATrain}{167,778}
\newcommand{\DSSQAVal}{117,082}
\newcommand{\DSSQATest}{151,256}
\newcommand{\DSHYETrain}{79,940}
\newcommand{\DSHYEVal}{45,886}
\newcommand{\DSHYETest}{47,411}
\newcommand{\DSBacterialChimTrain}{276,834}
\newcommand{\DSBacterialChimVal}{92,603}
\newcommand{\DSBacterialChimTest}{287,362}
\newcommand{\DSOxMetTrain}{135,926}
\newcommand{\DSOxMetVal}{60,511}
\newcommand{\DSOxMetTest}{181,479}
\newcommand{\DSKingdomsDeNovoCapTest}{4,926,232}

\newcommand{\DSKitchenSinkTotal}{5,791,021}

%% file: code/generated/tables/result_macros.tex
\newcommand{\ResDionMaxPeaksKingdoms}{0.607}
\newcommand{\ResDionMaxPeaksMSKB}{0.856}

%% file: code/generated/tables/result_macros_ablation.tex
\newcommand{\ResAblFrozenDual}{0.675}
\newcommand{\ResAblFrozenPure}{0.638}
\newcommand{\ResAblFrozenFree}{0.610}
\newcommand{\ResAblFrozenRandom}{0.582}

%% file: sections/abstract.tex
\begin{abstract}
We introduce a novel invariance for peptide tandem mass spectrometry data, unlocking self-supervised representation learning that improves de novo sequencing of peptides.
This invariance exploits the physical relationship between precursor properties (mass and charge) and fragment-ion evidence, without requiring peptide sequence labels.
We introduce \textbf{dIon}, which adapts the DINO framework with two latent prediction tasks, both recovering a clean teacher representation: one from a spectrum mixture, using the precursor as a selection query, and one from a partial spectrum with the precursor withheld. The first associates precursor information with fragment-ion evidence; the second prevents representational collapse onto that information alone.
Mechanistic probes support both effects, and ablations show that the full objective performs best. Under identical end-to-end training, dIon initialization improves de novo peptide precision over training from scratch by 5.5 and 8.4 percentage points on the held-out MassIVE-KB and Kingdoms test sets, and by 2.3 and 4.8 percentage points with a larger supervised training corpus. The resulting models surpass fully supervised state-of-the-art de novo sequencing models on the diverse, multi-species Kingdoms corpus under the same greedy-decoding protocol. Without peptide labels, dIon learns strong native peptide-similarity geometry compared with other learned models; with limited peptide-supervised adaptation, it achieves the best retrieval and pair-discrimination performance across all representation benchmarks.

\end{abstract}

%% file: sections/01_introduction.tex
\section{Introduction}
\label{sec:introduction}

Self-supervised learning has become a major approach for learning transferable
representations without task labels, using training objectives constructed from
the input data itself. Representative approaches include masked modeling
\citep{devlin2019bert,he2022mae}, contrastive learning across augmented views
\citep{chen2020simclr}, and non-contrastive methods that match or predict
learned targets across views
\citep{caron2021dino,baevski2022data2vec,assran2023ijepa}.
The self-supervised objective determines which structure in the data the model
must capture, making objective design especially important when moving to a
new modality.

For view-based methods, the construction of the views is part of the
self-supervised objective. In SimCLR, two augmented views of the same image
define a positive pair, and the choice of augmentations strongly affects the
learned representation \citep{chen2020simclr}. DINO trains a student to match
an exponential moving average (EMA) teacher across differently augmented global and local views of the same
image \citep{caron2021dino}. The views thus define which changes the representation should become
invariant to. For images, crops exploit the spatial structure of the pixel
grid. The self-distillation machinery itself (student, EMA teacher, stop-gradient) is modality agnostic; the views, however, should express an invariance appropriate for
that modality's data.

Peptide tandem mass spectrometry data has a very different structure.
A mass spectrum is an unordered set of peaks, each given by its measured
mass-to-charge ratio ($m/z$) and intensity. An intact peptide ion, the precursor, is measured in an MS1 scan and then
fragmented, and its fragment ions are measured in an MS2 scan, producing the
tandem MS/MS spectrum. Informative fragments may be absent,
unrelated or noise peaks may be present, and fragment intensities can differ
substantially between measurements, while the $m/z$ values of detected ions are
measured with high precision. An MS/MS spectrum is therefore a precise but
incomplete and variable observation of the fragmentation of a precursor with
measured mass and charge.
Public repositories such as PRIDE \citep{perezriverol2025pride} hold vast
quantities of MS/MS data, but only a minority of spectra carry peptide labels,
pseudo-labels from database search assigned where a spectrum confidently
matches a candidate peptide. Supervised training sees only this noisy,
non-randomly selected subset, whereas self-supervision can use every spectrum.

\begin{figure*}[t]
\centering
\begin{subfigure}[b]{0.44\textwidth}
  \centering
  \includegraphics[width=\linewidth]{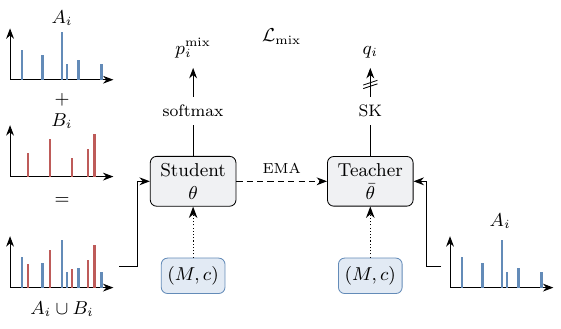}
  \caption{Precursor-conditioned mixture task}
\end{subfigure}\hspace{0.06\textwidth}
\begin{subfigure}[b]{0.44\textwidth}
  \centering
  \includegraphics[width=\linewidth]{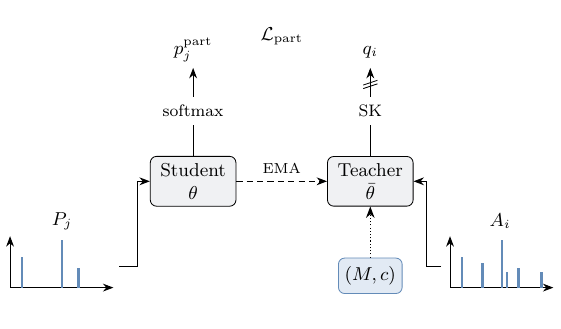}
  \caption{Precursor-withheld partial-view task}
\end{subfigure}
\caption{Dual latent-prediction objective. (a) The student sees the mixture
$A_i \cup B_i$ with the precursor $(M, c)$ of $A$ and matches the teacher
output $q_i$ for the clean view $A_i$ ($\mathcal{L}_{\mathrm{mix}}$). (b) The
student sees a partial view $P_j$ of $A$ with the precursor withheld and
matches the same target ($\mathcal{L}_{\mathrm{part}}$). The teacher is an EMA
of the student, and SK denotes Sinkhorn--Knopp balancing.}
\label{fig:method}
\end{figure*}

Masked modeling would seem a natural first choice. For peptide MS/MS, however,
the incompleteness and variability of spectra make the exact peaks and intensities
removed by masking a less appealing target for representation learning. We instead turn to latent prediction, where the model predicts a
learned target computed from another view rather than the masked observations
themselves
\citep{caron2021dino,baevski2022data2vec,lecun2022path,assran2023ijepa}.
Latent prediction still leaves the question of which relation should be
preserved across views. For modalities without an obvious analogue
of image crops or spatial masking, that relation can instead come from known
structure in the data-generating or measurement process. One way to turn such
a relation into a prediction task is to combine two data examples into a single
view, provide conditioning information associated with one of them, and ask the
model to recover that example's clean representation. Because the conditioning
information could itself become a shortcut, a second view withholds it and
presents only part of the example. We instantiate this idea
for peptide MS/MS using the physical relationship between precursor properties
and fragment-ion evidence, with no peptide sequence labels. We call this
\emph{fragmentation-based invariance}: the invariant relation is defined by
fragmentation physics, while the representation remains sensitive to the
fragmentation patterns that distinguish peptides.

We introduce \modelname{}, which adapts DINO self-distillation to peptide MS/MS
with two complementary latent-prediction tasks. Our main contributions are:
\begin{itemize}[topsep=3pt,itemsep=2pt,parsep=0pt,partopsep=0pt,leftmargin=1.5em]
   \item We introduce a self-supervised latent-prediction construction that
   mixes two examples and conditions recovery of a clean representation on
   information associated with one source, paired with partial views that
   withhold this information so that it cannot serve as a shortcut.

   \item We instantiate this construction for peptide MS/MS as
   \emph{fragmentation-based invariance}, through a
   \emph{precursor-conditioned mixture task} and a
   \emph{precursor-withheld partial-view task}, requiring no peptide sequence
   labels.

    \item We pretrain \modelname{} as a foundation encoder for peptide MS/MS. Its
    global spectrum representation serves peptide retrieval and pair
    discrimination, and its dense peak-token representations transfer to de
    novo peptide sequencing, which predicts a peptide's amino-acid sequence
    directly from its spectrum.

    \item Mechanistic probes support the intended behaviors of both
    latent-prediction tasks, and objective ablations show that the full dual
    objective performs best.

    \item \modelname{} initialization improves de novo peptide precision by
    2.3--8.4 percentage points under matched end-to-end training and surpasses
    the fully supervised Casanovo and InstaNovo models across all 70 species of the Kingdoms test set. To our knowledge, this is the first demonstration that self-supervised pretraining improves de novo sequencing over matched training from scratch.
    It also learns strong native peptide-similarity geometry compared with other
    learned models and improves label-efficient peptide-supervised metric
    learning.
\end{itemize}

%% file: sections/02_related_work.tex
\section{Related Work}
\label{sec:related-work}

View-based self-supervised learning spans contrastive methods
\citep{chen2020simclr,he2020moco} and non-contrastive ones
\citep{chen2021simsiam,grill2020byol,bardes2022vicreg,caron2021dino};
\modelname{} builds on DINO's teacher--student self-distillation. A second
family predicts learned targets from partial observations rather than
reconstructing the input: data2vec predicts contextualized teacher
representations from masked inputs \citep{baevski2022data2vec}, I-JEPA
predicts representations of target image regions from surrounding context
\citep{assran2023ijepa}, and Masked Siamese Networks match a masked view to a
target computed from the unmasked image \citep{assran2022msn}. Dense
downstream tasks have motivated explicit local objectives, from DenseCL's
local-feature contrast \citep{wang2021densecl} to iBOT's masked patch-token
prediction \citep{zhou2022ibot}, retained in DINOv2 \citep{oquab2024dinov2},
and Gram anchoring in DINOv3 \citep{simeoni2025dinov3}.

Mixing-based methods train on mixtures of inputs
\citep{lee2021imix,shen2022unmix,alex2025sslam}, but with targets that reflect
both sources. \modelname{} also mixes two examples, yet leaves the target
unblended: conditioning information associated with one source specifies whose
clean teacher output is wanted, and its partial views withhold that
information so it cannot serve as a shortcut.

For mass spectra, self-supervision has been explored mainly on small
molecules, combining masked-peak prediction with retention-order prediction
\citep{bushuiev2026dreams} or contrastive learning between masked views
\citep{zhang2024msbert}. Peptide-MS/MS representations have instead been
learned with peptide supervision: GLEAMS embeds spectra from database-search
assignments \citep{bittremieux2022gleams} and is our supervised reference in
Section~\ref{sec:geometry}, \citet{sanders2025foundation} transfer an encoder
pretrained by supervised de novo sequencing, and others learn from identified
peptide-spectrum matches or jointly with sequence prediction
\citep{altenburg2026yhydra,jin2024contranovo,zhao2026punifind}.
Closest to our setting is InstaNovo-FM, concurrent work that pretrains an
encoder to reconstruct masked regions of spectra without peptide-sequence
labels \citep{nieuwoudt2026instanovofm}; on their benchmark the authors report faster
convergence but no measurable de novo advantage over training the same
architecture directly. \modelname{} learns from teacher targets rather than
reconstructing the spectrum, and its initialization improves de novo
sequencing over matched training from scratch (Section~\ref{sec:denovo}).

De novo peptide sequencing has used deep learning since DeepNovo
\citep{tran2017deepnovo}, and modern Transformer models encode a spectrum into
contextualized peak representations that a peptide decoder attends over.
Casanovo established this formulation
\citep{yilmaz2022casanovo,yilmaz2024casanovo}, InstaNovo and InstaNovo+ follow
it under autoregressive and diffusion-based decoding \citep{eloff2025instanovo},
and Pairwise Attention biases the encoder's attention with pairwise mass
differences between peaks \citep{lapin2025pairwise}. That interface is the one
\modelname{}'s dense representations must be compatible with.

%% file: sections/03_method.tex
\section{Methods}
\label{sec:method}

\subsection{Spectrum representation and encoder}

A tandem mass spectrum is an unordered set of $N$ fragment-ion observations,
\begin{equation}
  S = \{(m_i, I_i)\}_{i=1}^{N},
  \label{eq:spectrum}
\end{equation}
where $m_i$ is the measured $m/z$ of peak $i$ and $I_i$ its intensity,
together with precursor information $(M, c)$, where $M$ is the neutral
precursor mass and $c$ the discrete precursor charge. Spectra are preprocessed
by restricting peaks to $m/z \le 2500$, retaining the $200$ most intense of
those, and normalizing intensities by the base peak.

Each peak is embedded from sinusoidal encodings of its $m/z$ and intensity, in
the manner commonly used to encode scalar quantities
such as position \citep{vaswani2017attention} and standard across Transformer
models for tandem mass spectra
\citep{yilmaz2024casanovo,eloff2025instanovo,yang2024pihelixnovo,lapin2025pairwise}
(Appendix~\ref{app:model_training_config}). Peaks are distinguished by their continuous $m/z$ rather than by a positional encoding over discrete indices. Precursor information is
encoded as a single token formed by summing a sinusoidal encoding of $M$ with
a learned embedding of the discrete charge $c$, and is prepended to the peak
tokens.

The spectrum encoder $f_\theta$ is a 15-block Transformer of width $1024$ with approximately
$127$M parameters, returning one contextualized representation $h_i$ per input
token. We pool these into a global spectrum representation $z$:
\begin{equation}
  \{h_i\}_{i=0}^{N} = f_\theta\bigl(S, (M, c)\bigr), \qquad
  z = \operatorname{Pool}\bigl(\{h_i\}_{i=0}^{N}\bigr).
  \label{eq:encoder}
\end{equation}
Here, $\operatorname{Pool}$ is a learned single-query cross-attention operation
applied after the final Transformer block. We call $h_i$ the dense peak-token representations and $z$ the global
representation, borrowing the distinction between global and dense features
drawn by DenseCL \citep{wang2021densecl}. Remaining architectural details are
given in Appendix~\ref{app:model_training_config}.

\subsection{Self-distillation}

\modelname{} is trained by self-distillation, following DINO
\citep{caron2021dino,oquab2024dinov2}. Each input is presented to the networks as several views. A view is a certain altered presentation of an input; in DINO augmented crops of an image. A student network must reproduce, from one view, what an exponential moving average (EMA) teacher produces from another view of the same input, without labels. Each
network pools its encoder output into a single representation $z_v$ per view
$v$, which a three-layer MLP head $g$ maps to a $K$-dimensional output.
Normalizing that output gives the assignment of the view, and the student's
assignment for one view must match the teacher's for another:
\begin{equation}
  p_v = \softmax\bigl(g_\theta(z_v) / \tau_s\bigr),
  \qquad
  q_v = \operatorname{SK}\bigl(g_{\bar\theta}(z_v) / \tau_t\bigr),
  \label{eq:assignments}
\end{equation}
where $\theta$ are the student weights and $\bar\theta$ their exponential
moving average, $\tau_s$ and $\tau_t$ are temperatures, and $\operatorname{SK}$
denotes Sinkhorn--Knopp balancing of the teacher assignments across the batch,
which DINOv2 \citep{oquab2024dinov2} also adopts from SwAV, in place of the
softmax centering used by DINO \citep{caron2021dino}. The student is trained
by minimizing the cross-entropy $H(q_v, p_{v'})$ between the teacher
assignment of one view and the student assignment of another.

The $K$ outputs are equivalently read as scores against $K$ learned vectors in
the head's last layer, the prototypes of SwAV and DINOv2
\citep{caron2020swav,oquab2024dinov2}. SK balancing and the low teacher
temperature together prevent output collapse, in which the teacher returns the
same assignment regardless of the input: balancing spreads a batch's assignment
mass across the $K$ outputs, while sharpening by $\tau_t$ keeps the assignments
from flattening to a uniform distribution, and either mechanism alone
encourages the failure the other prevents. No gradient flows through the
teacher branch, whose weights change only through the EMA update.

For \modelname{} we used $K = 16{,}384$. Arriving at schedules that train
stably on spectra required tuning $\tau_t$ and the EMA rate, monitored through
the entropy of the teacher assignments, the teacher--student KL, and the
across-batch variance of the student assignments
(Appendix~\ref{app:collapse}); the selected values and the remaining
hyperparameters are given in Appendix~\ref{app:ssl_objective}. Avoiding
constant output collapse is necessary but not sufficient, since a stable
configuration need not learn a useful representation.

\subsection{Fragmentation-based invariance}

The remaining question is which invariance to impose. It should be one that cannot be satisfied trivially, but require the model to capture structure useful for downstream transfer.
The views define this invariance through what they preserve, alter, or remove.
DINO constructs views using image transformations that exploit spatial structure, whereas tandem mass spectra are unordered sets of peaks. Peak $m/z$ values cannot be jittered freely because they are measured to within a few parts per million. Intensities vary more across measurements and can therefore be perturbed more naturally, but they carry substantially less information than peak $m/z$ values.
We instead derive the views from known structure in the measurement process, instantiated here by the physical relationship between precursor properties and fragment-ion evidence.

For each training example, we call the original spectrum the anchor $A$. All views are then constructed from $A$, with an additional mixture view adding peaks from a competing spectrum $B$.

\textbf{Clean teacher views.}
Clean views $A_i$ retain $95\%$ of the peaks of $A$, sampled independently.
They are encoded with the true precursor of $A$, and provide the
targets that every student view must recover.

\textbf{Precursor-conditioned mixture views.}
Each clean view $A_i$ is combined with the peaks of an independently sampled
competing spectrum $B_i$, giving the student input $A_i \cup B_i$. The two
peak sets are merged rather than concatenated: peaks coinciding within
$5$~ppm are combined into a single peak with summed intensity. A mixture
always combines the anchor with a single competing spectrum. The true anchor
precursor is retained, so it indicates which of the two peptides present in
the mixture should be represented. For this to be well posed, each mixture
view samples its competing spectrum uniformly from the spectra in the batch
whose neutral precursor mass differs from the anchor's by more than
$10$~ppm, which also excludes the same peptide observed at another charge.

\textbf{Precursor-withheld partial views.}
Partial views $P_j$ are formed by independently sampling $60\%$ of the peaks of
$A$ (swept over $\{30, 60\}\%$, Appendix~\ref{app:denovo_probe}). Their precursor token is replaced by a learned null embedding, so the
student receives no precursor information for these views. The peak
restriction is conceptually related to DINO's local crops
\citep{caron2021dino}: in both cases the student must match a target computed
from a more complete teacher view. The constructions themselves differ, since
a DINO local crop is a spatially contiguous region of an image, whereas a
partial spectrum is a random subset of peaks spanning the full $m/z$ range.
Withholding the precursor is a further restriction with no image-domain
counterpart, and its purpose is to deny the shortcut: a representation that
relied on precursor mass alone would be unable to recover the teacher target
here.

\subsection{Dual latent-prediction objective}

Figure~\ref{fig:method} summarizes the objective. The teacher encodes the
clean views, producing the targets. The student encodes the mixture views
and the partial views, and must recover those same targets: the mixture task
requires the student to focus on the signals corresponding to its input precursor while ignoring those of the interfering spectrum, while the partial-view task must recover the target from only a fraction of the anchor's peaks with no precursor at all.

Following Equation~\ref{eq:assignments}, we write $p^{\mathrm{mix}}_i$ for the
student assignment obtained from $A_i \cup B_i$, $p^{\mathrm{part}}_j$ for the
student assignment obtained from $P_j$ under the null precursor state, and
$q_i$ for the teacher assignment obtained from the clean view $A_i$.
In the mixture-task term, a mixture view with the anchor's precursor is matched to a clean teacher target with that same precursor,
\begin{equation}
  \mathcal{L}_{\mathrm{mix}}
    = \operatorname*{avg}_{i}\,
      H\bigl(q_i, p^{\mathrm{mix}}_{i}\bigr),
  \label{eq:loss-mix}
\end{equation}
whereas each partial view is matched to every clean target,
\begin{equation}
  \mathcal{L}_{\mathrm{part}}
    = \operatorname*{avg}_{i,\,j}\,
      H\bigl(q_i, p^{\mathrm{part}}_j\bigr),
  \label{eq:loss-loc}
\end{equation}
The two prediction tasks receive equal weight,
\begin{equation}
  \mathcal{L}_{\mathrm{\modelname{}}}
    = \tfrac{1}{2}\,\mathcal{L}_{\mathrm{mix}}
      + \tfrac{1}{2}\,\mathcal{L}_{\mathrm{part}},
  \label{eq:loss-dion}
\end{equation}
where
\begin{equation}
  H(q, p) = -q \log p
  \label{eq:cross-entropy}
\end{equation}
is the cross-entropy between a teacher assignment $q$ and a student assignment
$p$.

The objective is defined on $z$, and therefore the gradient update is computed through the learned query-pooling of equation 2. This contrasts with DINO, which generates $z$ from a \textsc{[cls]} token. Thus, the dense
representations $\{h_i\}$ are trained by the same objective, and \modelname{} uses no separate
dense-token loss of the kind
introduced by iBOT \citep{zhou2022ibot} and used in DINOv2
\citep{oquab2024dinov2}; Appendix~\ref{app:training_diagnostics} reports an
ablation that adds an iBOT-style dense loss term.

We use two views of each kind. Two is the most that fits in the memory of one
GH200 96\,GB accelerator at this model scale, and we did not sweep the count.
DINO reports that more views help, with diminishing returns
\citep{caron2021dino}. The head follows the DINOv2 design;
Appendix~\ref{app:ssl_objective} gives its configuration together with the
temperature and EMA schedules.

\subsection{Downstream transfer}

For de novo sequencing, the pretrained encoder is coupled to a freshly
initialized autoregressive Transformer decoder \citep{vaswani2017attention},
following the encoder--decoder formulation established for peptide MS/MS by
Casanovo \citep{yilmaz2022casanovo,yilmaz2024casanovo}: the decoder
cross-attends to the encoder's peak-token memory $\{h_i\}$, so de novo
sequencing consumes the dense representations rather than the pooled global
one, and both networks are optimized jointly under teacher-forced peptide-token
cross-entropy. Spectrum-level binary tasks instead apply a small multilayer
perceptron to $z$, and peptide-supervised metric learning applies a freshly
initialized projection head to $z$ trained with a supervised contrastive
objective \citep{khosla2020supcon}. Downstream architectures, optimization, and
decoding settings are given in Appendix~\ref{app:downstream_transfer} and
Appendix~\ref{app:denovo_dense}.

%% file: sections/04_experiments.tex
\section{Experiments and results}
\label{sec:experiments}
\label{sec:experimental-setup}
\label{sec:results}

\subsection{Pretraining and development protocol}
\label{sec:pretraining-protocol}

\modelname{} is pretrained without peptide labels on the \DatasetFull{bacterialssl}
of \citet{lee2022soilmetaproteomics} (\DSBacterialTrain{} spectra), holding out a run-disjoint subset of
\DSBacterialVal{} spectra for development (Appendix~\ref{app:datasets}). The
127M-parameter encoder is trained for 284 epochs on 64 GH200 accelerators in about
75 hours (Appendix~\ref{app:model_training_config}). Following DINO, we evaluate and release the EMA teacher. Checkpoints are selected
on development data and decoding settings are fixed in advance. Test splits are
used only for final evaluation (Appendix~\ref{app:model_selection}).

\subsection{De novo sequence transfer}
\label{sec:denovo}

All models, including the external baselines, are scored with one matcher,
which follows the criterion of DeepNovo \citep{tran2017deepnovo} as implemented
by Casanovo \citep{yilmaz2022casanovo}: treating a peptide as a sequence of
amino-acid tokens, a predicted token matches when its mass and its prefix mass
lie within 0.1 and 0.5~Da of the aligned reference token. Peptide precision at
full coverage is the fraction of all test spectra whose predicted peptide
matches at every token, with spectra lacking a
prediction counted as errors (Appendix~\ref{app:denovo_decoding}).

\textbf{Kingdoms evaluation.}
\modelname{} initialization improves peptide precision over training from
scratch in both training regimes (Table~\ref{tab:denovo-test}).
Fine-tuned on \DatasetFull{kitchensinkvfive} (\DatasetShort{kitchensinkvfive}), a
larger multi-source labeled corpus (Appendix~\ref{app:denovo_datasets}), it
also exceeds InstaNovo 1.2.2 and Casanovo 5.2.1 across all 70 species
(Table~\ref{tab:denovo-kingdoms-per-species}). When each model ranks its
predictions by its own confidence (Figure~\ref{fig:denovo-precision-coverage}),
\modelname{} is the most precise at every coverage level, tied with InstaNovo
only at the highest confidence.

\input{code/generated/tables/test_denovo_main}

\textbf{\DatasetShort{mskbfinal} evaluation.}
The same holds on \DatasetShort{mskbfinal}: \modelname{} initialization
improves over scratch in both training regimes, and every
\modelname{}-initialized model exceeds Casanovo 5.2.1
(Table~\ref{tab:denovo-test}). InstaNovo 1.2.2 is omitted there because it
was trained on MassIVE-KB v1, the source of this test set, and its splits are
unpublished.

\textbf{Larger peak cap.} For applications where prediction quality
matters more than compute, we also provide \modelname{}-initialized models
fine-tuned with a 1,000-peak cap. Although the encoder is pretrained with at
most 200 peaks per spectrum, the larger cap improves peptide precision further
on both benchmarks. Trained on \DatasetShort{kitchensinkvfive}, it reaches
\ResDionMaxPeaksKingdoms{} on Kingdoms and \ResDionMaxPeaksMSKB{} on
\DatasetShort{mskbfinal} (Tables~\ref{tab:denovo-kingdoms-capped-complete}
and~\ref{tab:denovo-mskb-complete}).

\textbf{Unlabeled sequencing on yeast MS2 data.}
We additionally evaluate on a public yeast run without peptide labels, using
proteome matching with a decoy-based false-discovery control
(Appendix~\ref{app:yeast_unlabeled}). On emitted predictions, the three models
fall within about one percentage point of each other
(Table~\ref{tab:denovo-yeast-unlabeled-complete}).

\subsection{Peptide-representation geometry}
\label{sec:geometry}

\textbf{Retrieval.}
Each spectrum queries the other spectra of its benchmark, and a retrieved
spectrum is relevant when it carries the query's peptide. Over the full
gallery, precursor mass and charge alone separate most candidates, so a high
\emph{broad} retrieval score does not necessarily show that a representation has learned
anything from the fragment ions. In practice these metadata are known for
every spectrum and are used to prefilter candidates, as GLEAMS does by charge
and precursor mass before clustering \citep{bittremieux2022gleams}. What a
representation must then contribute is to tell apart the remaining
near-isobaric peptides, which differ only in their fragment-ion patterns. We
therefore evaluate retrieval within such a prefilter, restricting the gallery
to spectra with the query's charge and a precursor $m/z$ within 10~ppm, and
report mean average precision (mAP) in this \emph{strict} setting
(Appendix~\ref{app:embedding_analysis}).

\textbf{Pair discrimination.}
Strict pairs are a fixed, balanced set of same-peptide and different-peptide
pairs under the same constraint, shared by all models. Retrieval asks whether
each query ranks its own peptide first. Pair ROC-AUC asks whether one
similarity scale separates same-peptide from different-peptide pairs across
the benchmark, as clustering with a single threshold requires.
Table~\ref{tab:gleams-matched-all} reports both strict metrics.

Without peptide labels, frozen \modelname{} exceeds InstaNovo-FM on every
benchmark and metric, and Casanovo 4.0, trained with full peptide supervision
for de novo sequencing, on four of the six (Table~\ref{tab:gleams-matched-all}).
GLEAMS \citep{bittremieux2022gleams}, trained supervised end to end
for peptide similarity on about 30 million labeled MassIVE-KB v1 spectra,
scores highest of these models and is the only one of them to match the
binned spectral angle.
Supervised metric learning changes the ranking. It fine-tunes the encoder with a
fresh projection head and a supervised contrastive loss over same-peptide
positives, using 1\% or 10\% of the training peptides
(Appendix~\ref{app:metric_learning}). With 10\% of the labels, \modelname{} initialization is best on every
benchmark and metric, and with 1\% it already exceeds GLEAMS on the Bacterial
and Kingdoms benchmarks. At both label budgets it matches or improves on
scratch initialization.
\input{code/generated/tables/validation_gleams_all}

\subsection{Mechanistic probes}

Two probes on the frozen encoder test the behaviours the dual objective was
designed to induce (Table~\ref{tab:validation-mechanism-summary}). In the
first, a fixed mixture $A \cup B$ is embedded under the precursor of $A$ or of
$B$, and a trial is correct when the representation lies closer to the clean
representation of the conditioned source than to that of the other. In the
second, two random 60\% peak subsets of $A$ are embedded with the precursor
withheld, and a view is correct when it lies closer to the clean
representation of $A$ than to that of a different-peptide spectrum.
Appendix~\ref{app:dual_view_interventions} gives other readouts.

\input{code/generated/tables/validation_mechanism_summary}

Conditioning on the anchor precursor selects the anchor almost always. The
competing precursor also selects its own source in the great majority of
trials, though less often, since it is capped at the anchor's peak count
(Equation~\ref{eq:competing-size}) and is never itself a target. The
goal of \modelname{} is to learn representations rather than to serve as a
final decomposition model. One aimed at decomposition would drop the cap and
add a symmetric term.

\subsection{Objective ablations}

We compare the dual objective of Section~\ref{sec:method} against two
baselines that each retain one half of it. All three are pretrained for only
100 epochs under otherwise identical settings. The pure mixture objective uses
only the mixture views. The mixture-free control uses only clean views and
swaps which views receive the precursor, so that the global views must infer
it from the fragments (Appendix~\ref{app:objective_ablations}). Neither
matches the dual objective.

The pure mixture objective reaches the highest broad retrieval mAP but the
weakest strict pair discrimination, and its strict pair AUROC falls as training
proceeds (Figure~\ref{fig:objective-ablation}). The same ordering holds on an
auxiliary probe of the tryptic C-terminal residue
(Appendix~\ref{app:objective_ablations}). This is the
representational collapse onto precursor information that the
precursor-withheld partial views are designed to prevent: without them, the
representation organizes spectra by the precursor metadata that broad
retrieval rewards (Section~\ref{sec:geometry}). The mixture-free control
avoids this collapse, and the dual objective, which adds the mixture views,
pulls clearly ahead on strict pairs in the second half of pretraining.

\newcommand{\abllegend}[2]{%
  \textcolor{#1}{\rule[0.28ex]{1.1em}{1.1pt}}\,\textcolor{#1}{#2}\;}
\begin{figure}[!htb]
\centering
{\footnotesize
 \abllegend{ablPure}{$\blacksquare$}Pure mixture \qquad
 \abllegend{ablFree}{$\blacktriangle$}Mixture-free \qquad
 \abllegend{ablDual}{$\bullet$}Dual objective\par}
\vspace{0.4em}
\begin{subfigure}[t]{0.40\linewidth}
  \centering
  \includegraphics[width=\linewidth,trim={0 0 0 23pt},clip]{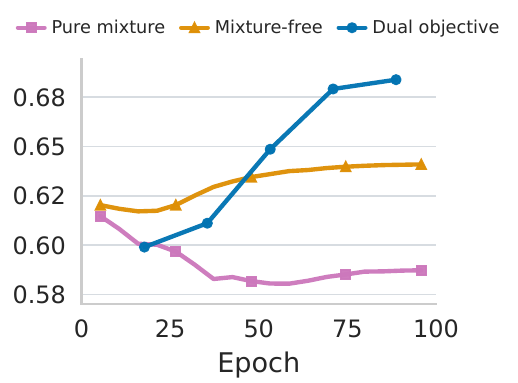}
  \caption{Strict pair AUROC}
\end{subfigure}\hspace{0.03\linewidth}
\begin{subfigure}[t]{0.40\linewidth}
  \centering
  \includegraphics[width=\linewidth,trim={0 0 0 23pt},clip]{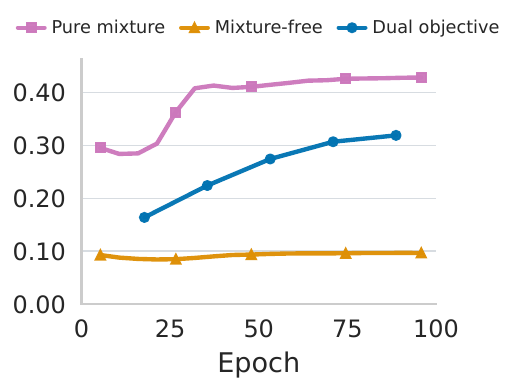}
  \caption{Broad retrieval mAP}
\end{subfigure}
\caption{\emph{Validation} probes during pretraining of the 100-epoch
ablation checkpoints. Strict is defined in Section~\ref{sec:geometry}.}
\label{fig:objective-ablation}
\end{figure}

\textbf{De novo transfer.} With the encoder frozen, the dual objective ranks
first in validation peptide precision (\ResAblFrozenDual{}), ahead of the pure mixture
objective (\ResAblFrozenPure{}), the mixture-free control (\ResAblFrozenFree{})
and a randomly initialized encoder (\ResAblFrozenRandom{}).
Appendix~\ref{app:objective_ablations} also reports end-to-end fine-tuning.

%% file: code/generated/tables/test_denovo_main.tex
\begin{table}[!htb]
\centering
\small
\caption{De novo sequencing on the \DatasetShort{kingdomsdenovo} (all 70 species) and \DatasetShort{mskbfinal} test sets. Within each training regime, scratch and \modelname{} initialization use an identical end-to-end training protocol.}
\label{tab:denovo-test}
\begin{threeparttable}
\begin{tabular}{@{}lS[table-format=1.3]S[table-format=1.3]@{}}
\toprule
 & \multicolumn{2}{c}{Peptide precision} \\
\cmidrule(lr){2-3}
{Training / initialization} & {Kingdoms} & {MSKB-final} \\
\midrule
\multicolumn{3}{@{}l}{\itshape MSKB-final training} \\
Scratch, end-to-end & 0.449 & 0.747 \\
\modelname{} init., end-to-end & 0.533 & 0.802 \\
\addlinespace
\multicolumn{3}{@{}l}{\itshape \DatasetFull{kitchensinkvfive} training} \\
Scratch, end-to-end & 0.554 & 0.819 \\
\modelname{} init., end-to-end & \textbf{0.602} & \textbf{0.842} \\
\addlinespace
\multicolumn{3}{@{}l}{\itshape External reference} \\
Casanovo 5.2.1 & 0.492 & 0.783 \\
InstaNovo 1.2.2 & 0.570 & \multicolumn{1}{c}{\textemdash} \\
\bottomrule
\end{tabular}
\end{threeparttable}
\end{table}

%% file: code/generated/tables/validation_gleams_all.tex
\begin{table*}[!htb]
\centering
\small
\caption{Representation geometry across test benchmarks (Kingdoms: all 70 species).}
\label{tab:gleams-matched-all}
\begin{threeparttable}
\begin{tabular}{@{}lS[table-format=1.3]S[table-format=1.3]S[table-format=1.3]S[table-format=1.3]S[table-format=1.3]S[table-format=1.3]@{}}
\toprule
 & \multicolumn{2}{c}{Bacterial} & \multicolumn{2}{c}{NineSpecies} & \multicolumn{2}{c}{Kingdoms} \\
\cmidrule(lr){2-3}
\cmidrule(lr){4-5}
\cmidrule(lr){6-7}
 & \multicolumn{1}{c}{Retrieval} & \multicolumn{1}{c}{Pairs} & \multicolumn{1}{c}{Retrieval} & \multicolumn{1}{c}{Pairs} & \multicolumn{1}{c}{Retrieval} & \multicolumn{1}{c}{Pairs} \\
{Method} & {mAP} & {AUROC} & {mAP} & {AUROC} & {mAP} & {AUROC} \\
\midrule
Binned spectral angle & 0.915 & 0.943 & 0.975 & 0.954 & 0.905 & 0.871 \\
\addlinespace
\multicolumn{7}{@{}l}{\itshape Peptide-supervised} \\
Casanovo 4.0 & 0.881 & 0.898 & 0.932 & 0.859 & 0.838 & 0.746 \\
GLEAMS & 0.916 & 0.948 & 0.976 & 0.962 & 0.903 & 0.876 \\
\addlinespace
\multicolumn{7}{@{}l}{\itshape Self-supervised} \\
InstaNovo-FM & 0.847 & 0.864 & 0.874 & 0.763 & 0.805 & 0.708 \\
\modelname{} & 0.895 & 0.928 & 0.926 & 0.865 & 0.821 & 0.748 \\
\addlinespace[0.6em]
\multicolumn{7}{@{}l}{\itshape Peptide-supervised contrastive} \\
Scratch init. (1\%) & 0.909 & 0.934 & 0.972 & 0.946 & 0.906 & 0.873 \\
\modelname{} init. (1\%) & 0.922 & 0.953 & 0.972 & 0.953 & 0.911 & 0.887 \\
Scratch init. (10\%) & 0.928 & 0.960 & \textbf{0.978} & 0.957 & 0.930 & 0.902 \\
\modelname{} init. (10\%) & \textbf{0.933} & \textbf{0.968} & \textbf{0.978} & \textbf{0.965} & \textbf{0.940} & \textbf{0.926} \\
\bottomrule
\end{tabular}
\end{threeparttable}
\end{table*}

%% file: code/generated/tables/validation_mechanism_summary.tex
\begin{table}[!htb]
\centering
\small
\caption{Source recovery by \modelname{} on the \DatasetShort{ninespecies} \emph{validation} set, by cosine similarity of the global representation.}
\label{tab:validation-mechanism-summary}
\begin{threeparttable}
\begin{tabular}{@{}lS[table-format=1.3]S[table-format=1.3]S[table-format=1.3]@{}}
\toprule
 & \multicolumn{3}{c}{Correct-source accuracy} \\
\cmidrule(lr){2-4}
{Probe} & {First} & {Second} & {Both} \\
\midrule
Mixture $A \cup B$, precursor of $A$\,/\,$B$ & 0.997 & 0.929 & 0.926 \\
Partial views $P_1$\,/\,$P_2$ of $A$, null precursor & 0.987 & 0.984 & 0.978 \\
\bottomrule
\end{tabular}
\end{threeparttable}
\end{table}

%% file: sections/05_discussion.tex
\section{Discussion}
\label{sec:discussion}

The self-distillation machinery of DINO transfers to peptide tandem mass
spectra unchanged; what does not transfer is the view construction built on
image geometry. The fragmentation-based invariance is that adaptation: it
derives the invariant relation from the structure of the
measurement, and the dual objective imposes it without labels. The failure mode it avoids is not necessarily specific to this modality. Conditioning information
supplied alongside an observation is a shortcut, and an objective that always
provides it will be solved using it, much as SimCLR needs color distortion
because crops of one image can be matched by their color histograms alone
\citep{chen2020simclr}. Withholding the precursor in a second view family
denies the analogous shortcut here, and neither branch alone reproduces the
full result (Figure~\ref{fig:objective-ablation}).

Unlike masked reconstruction, which gave no measurable de novo advantage in
InstaNovo-FM \citep{nieuwoudt2026instanovofm}, predicting learned teacher
targets improves de novo sequencing over matched training from scratch,
consistent with our argument that reconstructing incomplete and variable peaks
is a poor pretraining target for peptide MS/MS. Although the objective is defined only on the pooled global representation,
the dense per-token representations benefit a decoder that attends over them,
and an explicit token-level objective of the kind used by iBOT and DINOv2 did
not help (Appendix~\ref{app:training_diagnostics}). We attribute this to the
pooling and the partial views. Because $z$ is pooled from the token outputs by
learned-query attention rather than read from a \textsc{[cls]} token, the
objective acts on the tokens themselves, and because a partial view holds only
a random subset of peaks, the tokens must stay informative without the full
spectrum.
Gram anchoring, which DINOv3 introduced against dense-feature degradation
\citep{simeoni2025dinov3}, helps here too: anchoring to a 30\% partial-view
checkpoint with stronger dense probes improves frozen but not fine-tuned de
novo transfer (Appendix~\ref{app:denovo_refinement}). The dense information is
therefore present without anchoring and is recovered when the encoder is
adapted.

Frozen, the training-free binned spectral angle outperforms every learned
representation not trained for peptide similarity, including the
peptide-supervised Casanovo encoder (Table~\ref{tab:gleams-matched-all}). With
supervised metric learning on 10\% of the labels, \modelname{} initialization
outperforms every method on every benchmark and metric, GLEAMS included. On three auxiliary
spectrum-level tasks, fine-tuning from \modelname{} likewise exceeds training
from scratch on all three, and every other evaluated representation on spectrum
quality and methionine oxidation. Only on these auxiliary tasks does its frozen
representation trail both Casanovo and InstaNovo-FM, and on spectrum quality
only negligibly (Appendix~\ref{app:downstream_tasks}).

A limitation we share with the field is that our evaluations score predictions
against peptide labels assigned by a search engine, so they inherit its errors
and selection bias. Only pretraining is free of these labels. The
one exception is the yeast evaluation, which instead checks whether each
predicted peptide occurs in the yeast proteome
(Appendix~\ref{app:yeast_unlabeled}). Pretraining at
this scale is also costly (Appendix~\ref{app:compute}). Although not trained as a decomposition
model, \modelname{} can decompose a mixture given each component's precursor,
so it could potentially be used for data-independent acquisition (DIA), where
wide isolation windows co-fragment several peptides. Appendix~\ref{app:dia_oracle} reports a pilot on real DIA
spectra.

Because the objective needs no labels, pretraining can in princple draw on every acquired
spectrum, including the majority that search engines leave unassigned.

%% file: sections/06_conclusion.tex
\section{Conclusion}
\label{sec:conclusion}

View-based self-distillation can be carried to a new modality by deriving its
views from a relation that the measurement itself supplies. For
peptide tandem mass spectra that relation is the physical link between a
precursor and the fragments it produces, and imposing it needs no peptide
labels. Under identical end-to-end training, \modelname{} initialization
improves de novo peptide precision over training from scratch by between $2.3$
and $8.4$ percentage points across both test sets and both supervised training
corpora, and exceeds the supervised state of the art on
\DatasetShort{kingdomsdenovo}. Without labels it learns strong native
peptide-similarity geometry compared with other learned models, and as an
initialization for peptide-supervised metric learning it exceeds training from scratch at both label budgets and, with 10\% of the labels,
outperforms every compared method. With \modelname{} pretraining, useful global and dense representations emerge
from unlabeled spectra alone.

%% file: appendix/A_datasets.tex
\section{Datasets, splits, and provenance}
\label{app:datasets}

\subsection{Overview of datasets and roles}

Table~\ref{tab:dataset-overview} lists every dataset with its role, split
sizes and split criterion.

\input{tables/dataset_overview}

\subsection{SSL pretraining and development corpus}

\modelname{} is pretrained on the \DatasetFull{bacterialssl}
\citep{lee2022soilmetaproteomics}, which covers the proteomes of 51 bacterial
organisms and comprises \DSBacterialTrain{} charge-1--10 training spectra. The held-out
\DSBacterialVal{}-spectrum subset is used for development. The underlying
peptide-spectrum matches provide corpus construction metadata, but the SSL
objective receives only peaks and precursor information: peptide sequence,
identity, PSM assignment, and fragment annotation are withheld.

\subsection{Native-geometry evaluation datasets}

We construct the bacterial native-geometry benchmark from
\DatasetShort{bacterialssl} validation spectra for development and
\DatasetShort{bacterialbenchmark} spectra for held-out testing. The derived test
retrieval population contains \DSBacterialTest{} spectra and is run-disjoint
from the \DatasetShort{bacterialssl} pretraining data, and test spectra whose exact modified peptide sequence occurs
in \DatasetShort{bacterialssl} training or validation are removed. The two corpora were
released together as the training and test sets of a de novo model for soil
metaproteomics \citep{lee2022soilmetaproteomics}, and they share no organism:
\DatasetShort{bacterialbenchmark} covers four bacterial species absent from the 51
organisms of \DatasetShort{bacterialssl}.

\begin{table}[!htb]
\centering
\small
\caption{Held-out retrieval spectra. The charge-2--4 column is the matched
subset used for external-reference comparisons.}
\label{tab:native-dataset-splits}
\begin{tabular}{@{}lrrp{0.43\linewidth}@{}}
\toprule
Corpus & Full-charge & Charge-2--4 & Split definition \\
\midrule
\DatasetShort{bacterialbenchmark} & \DSBacterialTest & \DSBacterialTestChargeTwoToFour & PXD010613 runs are held out from PXD010000 development data; exact modified sequences overlapping PXD010000 train/validation are removed. \\
\DatasetShort{ninespecies} & \DSNineSpeciesTest & \DSNineSpeciesTestChargeTwoToFour & Reserved fixed paper test; development excludes normalized peptide backbones appearing in this test. \\
\DatasetShort{kingdoms} & \DSKingdomsTest & \DSKingdomsTestChargeTwoToFour & For 10 multi-batch species, validation holds out one deterministic acquisition batch; test uses remaining files plus all other species. Run-disjoint where batch structure exists, not universally project-disjoint. \\
\bottomrule
\end{tabular}
\end{table}

\DatasetFull{ninespecies} (MassIVE accession MSV000090982;
\citealp{yilmaz2024casanovo}) evaluates nine organisms using separate development
and reserved evaluation populations; the latter contains
\DSNineSpeciesTest{} retrieval spectra. Peptide backbones assigned to the
reserved population are excluded from development using uppercase amino-acid
normalization. The \DatasetFull{kingdoms} \citep{muller2020kingdoms} provides a larger cross-species setting.
Its retrieval, pair and de novo test sets all cover the same 70 species, with
\DSKingdomsTest{} retrieval spectra in the held-out population. Its
validation and test sets are separated by acquisition run but are intentionally
not peptidoform-disjoint.

Retrieval and static-pair evaluation are separate views of each corpus.
Mass-controlled comparisons require equal precursor charge and precursor mass
within 10~ppm. GLEAMS comparisons use deterministic charge-2--4 subsets on
which every method has an embedding; these are reported separately from the
charge-complete populations. The restriction is small: the charge-2--4 cohort
retains $96.3\%$ of \DatasetShort{bacterialbenchmark}, $99.7\%$ of
\DatasetShort{ninespecies}, and $99.4\%$ of \DatasetShort{kingdoms},
or $99.1\%$ of the three benchmarks combined.%

\subsection{Supervised metric-learning data}

Metric learning uses peptide identities from the bacterial corpus after SSL
pretraining. Label budgets are sampled at the peptide-identity level, with all
eligible spectra for a selected identity retained; the 1\% subset is nested
within the 10\% subset. Evaluation populations and the supervised contrastive
head are held fixed across scratch and \modelname{} initialization, isolating
the contribution of SSL pretraining.

\subsection{De novo datasets and split construction}
\label{app:denovo_datasets}

The \DatasetFull{mskbfinal} is the released two-million-PSM tryptic/non-tryptic
MassIVE-KB corpus used to train Casanovo 4.2 and 5.0
\citep{melendez2024enzymebias,melendez2024mskbfinaldata}. Tryptic examples
originate from \DatasetFull{massivekbvone}, while the multi-enzyme component comes from the
corresponding \DatasetFull{massivekbvtwo} subset. We preserve the 200,000-spectrum
Casanovo test split and draw a seed-42 validation subset of 10,004 spectra from
the remaining data. Sampling is grouped by canonical peptidoform without
charge, preventing train--validation peptidoform overlap.

The \DatasetFull{kitchensinkvfive} (\DatasetShort{kitchensinkvfive}) corpus
broadens the peptide, enzyme and charge coverage of \DatasetShort{mskbfinal}
with \DSKitchenSinkTotal{} labeled spectra from eight sources
(Table~\ref{tab:dnl-sources}). It contains all of \DatasetShort{mskbfinal},
with its test split kept as test and a validation subset carved out of its
training split. Synthetic human peptides come from ProteomeTools Parts I--III
\citep{zolg2017proteometools,gessulat2019prosit}, where Part III adds HLA
class I and II and non-tryptic peptides. Their PSMs are kept at a posterior
error probability of at most 0.01, with reverse and contaminant hits removed.
\DatasetFull{massivekbvtwo} \citep{wang2018massivekb} contributes PSMs at 1\%
FDR from public MassIVE datasets, with chimeric spectra removed. The
\DatasetFull{bacterialssl} contributes its database-search PSMs, and the
non-enzymatic Casanovo training set \citep{yilmaz2024casanovo} contributes
MassIVE-KB and PROSPECT PSMs balanced over C-terminal residues. A high-charge
collection raises the share of precursors with charge 5 or more. It combines
PXD053100, PXD018061, PXD007648 and PXD034980
\citep{lim2025dsrna,kim2020tent4,gorshkov2019corona,li2024fkbpl} with
additional high-charge spectra that will be released at a later date. The added sources are subsampled with per-charge quotas and,
for all but the high-charge collection, a cap of 3,000 spectra per run. The
corpus contains no Kingdoms data, and it shares no acquisition run with the
\DatasetShort{kingdomsdenovo} test set.

Peptidoforms outside \DatasetShort{mskbfinal} are split 90/5/5 into training,
validation and test. A global test--validation--train precedence rule then
assigns each canonical peptidoform to one split, and exact peak-list duplicates
are collapsed. We checked the finished corpus for peptidoform overlap: its
splits share no canonical peptidoform, and no \DatasetShort{mskbfinal} test
peptidoform occurs in its training or validation splits. The duplicate
collapse removes eight spectra from the \DatasetShort{mskbfinal} test
component, so the \DatasetShort{kitchensinkvfive} test set is closely related
to, but not identical to, the released \DatasetShort{mskbfinal} test split.

\input{code/generated/tables/dataset_dnl_v1_sources}

\textbf{Kingdoms capped de novo evaluation.}
We derive the \DatasetFull{kingdomsdenovo} set by applying a deterministic
species-stratified cap to the existing Kingdoms
run-disjoint test split. The original test population has 14,519,567 spectra
across 70 species; the cap retains 4,926,232 spectra (33.93\%). Each species
contributes at most 100,000 spectra: 33 species are capped and smaller species
are retained in full. Within species, each spectrum receives a pseudo-random
number, the BLAKE2b hash of its species, peak file, scan ID and source-row index
with seed 20260915, and the 100,000 spectra with the smallest numbers are
retained. This gives a uniform random sample that is deterministic and
independent of row order.
Validation is copied unchanged.
The resulting cohort remains run-disjoint and is not peptidoform-disjoint.

Casanovo 4.0 and MSKB-final serve different purposes in this study. The
Casanovo 4.0 encoder is the representation reference examined in Casanovo
Foundation \citep{yilmaz2024casanovo,sanders2025foundation}. It was trained on
30 million \DatasetFull{massivekbvone} PSMs, retaining up to 100 PSMs per
peptidoform--charge precursor \citep{bittremieux2025massivekb30m}. Casanovo 4.0
also provides a separately trained non-enzymatic model based on \DatasetShort{massivekbvone} and
\DatasetFull{prospect}. By contrast, our matched supervised de novo experiments train on
\DatasetShort{mskbfinal}, the smaller mixed-enzyme corpus associated with Casanovo 4.2 and
5.0, and on \DatasetShort{kitchensinkvfive}.

\subsection{Auxiliary downstream datasets}

We derive the spectrum-quality task from a 19-project human Orbitrap HCD
collection, using a class-balanced charge-1--10 subset. Chimericity labels are
derived from FragPipe/MSFragger wide-window searches at 1\% PSM FDR: the
primary mixture task uses the \DatasetFull{hye}, while a second task uses the
\DatasetFull{bacterialbenchmark} with species-disjoint splits. From the same
bacterial source corpus, we construct an oxidized-methionine task on
methionine-containing peptides.

%% file: tables/dataset_overview.tex
\begin{table*}[!ht]
\centering
\footnotesize
\setlength{\tabcolsep}{4pt}
\caption{Datasets used in the principal experiments, with spectrum counts per
split and the split criterion.}
\label{tab:dataset-overview}
\begin{tabular}{@{}llrrr>{\raggedright\arraybackslash}p{0.27\textwidth}@{}}
\toprule
Dataset & Use & Train & Validation & Test & Split \\
\midrule
PXD010000 & SSL pretraining & \DSBacterialTrain & \DSBacterialVal & \textemdash & Acquisition project / run \\
PXD010613 & Retrieval & \textemdash & \DSBacterialGeomVal & \DSBacterialTest & Run held out; exact sequence filtered; development from PXD010000 validation \\
NineSpecies & Retrieval & \textemdash & \DSNineSpeciesVal & \DSNineSpeciesTest & Reserved test; test peptides excluded from development \\
Kingdoms & Retrieval & \textemdash & \DSKingdomsVal & \DSKingdomsTest & Acquisition batch; run-disjoint where available; 70 species \\
\midrule
MSKB-final & De novo & \DSMSKBTrain & \DSMSKBVal & \DSMSKBTest & Casanovo split; canonical peptidoform \\
\DatasetShort{kitchensinkvfive} & De novo training & \DSKitchenSinkTrain & \DSKitchenSinkVal & \DSKitchenSinkTest & Global peptidoform separation \\
Kingdoms de novo & De novo test & \textemdash & \textemdash & \DSKingdomsDeNovoCapTest & Run-disjoint; all 70 species, at most 100k spectra each \\
PXD026806 & Label-free proteome & \textemdash & \textemdash & 39,460 & Single yeast acquisition run \\
\midrule
SQA & Spectrum quality & \DSSQATrain & \DSSQAVal & \DSSQATest & Released project split \\
HYE & Chimericity & \DSHYETrain & \DSHYEVal & \DSHYETest & HYE partition \\
Bacterial chim. & Chimericity transfer & \DSBacterialChimTrain & \DSBacterialChimVal & \DSBacterialChimTest & Bacterial species \\
OxMet & Oxidized Met & \DSOxMetTrain & \DSOxMetVal & \DSOxMetTest & Bacterial species \\
\bottomrule
\end{tabular}
\end{table*}

%% file: code/generated/tables/dataset_dnl_v1_sources.tex
\begin{table}[!htb]
\centering
\small
\caption{Sources of \DatasetShort{kitchensinkvfive}, with spectra per split and precursor-charge range.}
\label{tab:dnl-sources}
\begin{tabular}{@{}lrrrc@{}}
\toprule
Source & Train & Validation & Test & Charges \\
\midrule
\DatasetShort{mskbfinal} & 1,974,956 & 25,002 & 199,992 & 2--5 \\
\DatasetFull{bacterialssl} & 863,005 & 48,273 & 48,073 & 1--10 \\
ProteomeTools Part III (PXD021013) & 788,303 & 37,086 & 37,426 & 1--7 \\
ProteomeTools Part I (PXD004732) & 508,466 & 12,662 & 12,939 & 2--7 \\
\DatasetShort{massivekbvtwo} & 416,182 & 17,891 & 17,472 & 1--8 \\
ProteomeTools Part II (PXD010595) & 382,417 & 16,881 & 16,576 & 2--6 \\
Casanovo non-enzymatic set & 223,952 & 6,838 & 7,044 & 1--7 \\
High-charge collection & 118,186 & 5,507 & 5,892 & 2--7 \\
\midrule
Total & 5,275,467 & 170,140 & 345,414 & 1--10 \\
\bottomrule
\end{tabular}
\end{table}

%% file: appendix/B_architecture_training.tex
\section{Architecture and training configuration}
\label{app:architecture_training}

\subsection{Model and SSL configuration}
\label{app:model_training_config}

The \modelname{} encoder is a 15-block Transformer with model dimension
$d = 1024$ and approximately 127.3M parameters. Layer normalization
is applied after each residual branch, and dropout of $0.15$ is applied to the attention
and feed-forward residual-branch outputs, with no dropout on the attention
weights themselves. Table~\ref{tab:encoder-architecture} summarizes the
configuration.

\begin{table}[!htb]
\centering
\small
\caption{\modelname{} spectrum encoder configuration.}
\label{tab:encoder-architecture}
\begin{tabular}{@{}ll@{}}
\toprule
Component & Setting \\
\midrule
Transformer blocks & 15 \\
Model dimension $d$ & 1024 \\
Attention heads & 8 (128 dimensions per head) \\
Feed-forward network & Linear(1024, 2048) $\rightarrow$ ReLU $\rightarrow$ Linear(2048, 1024) \\
Normalization & Post-residual LayerNorm (attention and FFN branches) \\
Dropout & 0.15 (attention and FFN residual outputs) \\
Parameters & $\approx$127.3M \\
\addlinespace
$m/z$ encoding & Four components, 256-D sinusoidal each (wavelengths 1--500); 1024 total \\
Intensity encoding & 256-D sinusoidal (wavelengths $10^{-6}$--$1$) \\
Peak projection & Bias-free Linear(1280, 1024) \\
Precursor mass encoding & 1024-D sinusoidal (wavelengths $10^{-3}$--$10^{4}$) \\
Charge embedding & Learned Embedding(11, 1024), charges 0--10 \\
Precursor token & Sum of mass encoding and charge embedding, prepended \\
Collision energy & Not used \\
\textsc{[cls]} token & None \\
\addlinespace
Global pooling & One learned 1024-D query, 8-head cross-attention \\
Pooling scope & Non-padding encoder tokens \\
\bottomrule
\end{tabular}
\end{table}

\subsubsection{Peak preprocessing}

Every spectrum supplied to the model is preprocessed identically: peaks are
restricted to $0 \le m/z \le 2500$; if more than 200 peaks remain, the 200
most intense are retained and intensities are base-peak normalized,
\begin{equation}
  I_i' = \frac{I_i}{\max_j I_j + 10^{-12}},
  \label{eq:base-peak-norm}
\end{equation}
where the maximum runs over the retained peaks. Spectra are then
zero-padded within each batch and supplied with a padding mask. This
preprocessing is shared
by \modelname{}, the scratch and randomly initialized controls, the
metric-learning fine-tunes, the auxiliary-task models, and the in-house de
novo models. External reference models use their own released preprocessing.
The de novo models fine-tuned with a 1,000-peak cap retain the 1,000 most
intense peaks instead.

\subsubsection{Peak and precursor encoding}

Peak features concatenate sinusoidal encodings of $m/z$ and base-peak-
normalized intensity. The $m/z$ value $m_i$ is decomposed into four numeric
components: the 100-Da bin, the integer remainder within that bin, and the
first and second pairs of fractional decimal digits,
\begin{equation}
  d_1 = \Bigl\lfloor \tfrac{m_i}{100} \Bigr\rfloor, \quad
  d_2 = \lfloor m_i \rfloor - 100\,d_1, \quad
  d_3 = \Bigl\lfloor \tfrac{r_i}{100} \Bigr\rfloor, \quad
  d_4 = r_i - 100\,d_3,
  \label{eq:mz-subdivision}
\end{equation}
with $r_i = \operatorname{round}\bigl(10^{4}\,(m_i - \lfloor m_i \rfloor)\bigr)$.
Each component receives a 256-dimensional sinusoidal encoding $\phi_{256}$
over wavelengths $1$--$500$, giving a 1024-dimensional $m/z$ representation,
and the base-peak-normalized intensity $I_i'$ receives a separate
256-dimensional encoding over wavelengths $10^{-6}$--$1$. The resulting
1280-dimensional peak feature is mapped to the model width by a bias-free
linear projection,
\begin{equation}
  p_i = W_{\mathrm{peak}}
    \bigl[\phi_{256}(d_1);\, \phi_{256}(d_2);\, \phi_{256}(d_3);\,
          \phi_{256}(d_4);\, \phi_{256}(I_i')\bigr],
  \qquad W_{\mathrm{peak}} \in \mathbb{R}^{1024 \times 1280}.
  \label{eq:peak-projection}
\end{equation}
The projected 1024-dimensional peak tokens enter the 15 Transformer blocks
directly.

The precursor token is
\begin{equation}
  c_{(M,c)} = \phi_{1024}(M) + \mathrm{Emb}(c) \in \mathbb{R}^{1024},
  \label{eq:precursor-token}
\end{equation}
where the neutral precursor mass $M$ receives a 1024-dimensional sinusoidal
encoding over wavelengths $10^{-3}$--$10^{4}$ and $\mathrm{Emb}$ is a learned
$\mathrm{Embedding}(11, 1024)$ over charges $0$--$10$. No sequence-position
encoding is added; the model relies on the physical $m/z$ encoding rather than
peak-list position. The null precursor embedding used by the precursor-withheld partial
views is obtained by setting the neutral mass input to zero and the charge to
the reserved index $0$; charge $0$ never occurs for a real spectrum, so its
row of the charge embedding is learned exclusively for this state.

The encoder exports two representations. The dense representations
$\{h_i\}_{i=1}^{N}$ are the contextualized peak-token outputs and form the
encoder memory consumed by de novo decoders. The global representation is
produced by learned-query pooling with a single learned query
$u \in \mathbb{R}^{1024}$ and 8-head cross-attention over the non-padding
encoder tokens,
\begin{equation}
  z = \operatorname{MHA}\bigl(u,\, \{h_i\}_{i=0}^{N},\, \{h_i\}_{i=0}^{N}\bigr)
    \in \mathbb{R}^{1024}.
  \label{eq:pooling}
\end{equation}

\subsubsection{View construction}

Table~\ref{tab:view-configuration} gives the view multiplicities and mixture
settings of the selected configuration. Each anchor spectrum yields two clean
teacher global crops, two precursor-conditioned mixture views built from the
same two crops, and two precursor-withheld partial views. The replicate count of
two per kind was not swept or ablated: it is the largest multi-view
configuration that fits in the memory of a single GH200 96\,GB accelerator at
this model scale, and DINO reports that additional views improve
representations with diminishing returns \citep{caron2021dino}.

\begin{table}[!htb]
\centering
\small
\caption{View construction and objective weighting for the selected
\modelname{} configuration. Peak fractions are uniform random subsets of the
anchor spectrum.}
\label{tab:view-configuration}
\begin{tabular}{@{}lll@{}}
\toprule
View & Count & Construction \\
\midrule
Clean teacher global & 2 & 95\% anchor subset, true precursor \\
Mixture view & 2 & 95\% anchor subset $+$ equal-size competing subset, true precursor \\
Partial view & 2 & 60\% anchor subset, reserved null precursor (mass 0, charge 0) \\
\midrule
\multicolumn{3}{@{}l}{\itshape Mixture settings} \\
Mixing strength & \multicolumn{2}{l}{1.0 (competing subset matches anchor subset peak count)} \\
Competing spectrum & \multicolumn{2}{l}{Independently sampled per view from the same batch} \\
Eligibility & \multicolumn{2}{l}{$>10$~ppm separation under same-charge precursor $m/z$ and neutral mass} \\
Peak merging & \multicolumn{2}{l}{Within 5~ppm after mixing; summed intensity, intensity-weighted $m/z$} \\
Intensity renormalization & \multicolumn{2}{l}{None after mixing} \\
\midrule
\multicolumn{3}{@{}l}{\itshape Objective weighting} \\
Mixture task & \multicolumn{2}{l}{0.5} \\
Partial-view task & \multicolumn{2}{l}{0.5} \\
Prototypes $K$ & \multicolumn{2}{l}{16,384} \\
\bottomrule
\end{tabular}
\end{table}

\textbf{Mixture construction.}
For an anchor spectrum $A$, a clean crop $A_i$ retaining $95\%$ of its peaks
is formed and an eligible in-batch competing spectrum $B$ is selected. The
competing contribution is explicitly size-matched to the anchor crop:
\begin{equation}
  n_B = \min\bigl(n_A,\, N_B\bigr)
  \label{eq:competing-size}
\end{equation}
peaks are sampled uniformly without replacement from $B$, where $n_A = |A_i|$
and $N_B$ is the number of peaks available in $B$. At the selected mixing
strength of $1.0$ this is exactly one competing peak per anchor-crop peak
whenever $B$ holds at least $n_A$ peaks, which prevents unusually dense
competing spectra from dominating a mixture through peak count alone. The
mixed view is the ppm-aware union
\begin{equation}
  M_i = \operatorname{merge}_{5\,\mathrm{ppm}}\bigl(A_i \cup B_i\bigr),
  \label{eq:mixture-union}
\end{equation}
formed by concatenating the two peak sets, sorting by $m/z$, and merging
connected adjacent peaks within $5$~ppm. Each merged group takes summed
intensity and intensity-weighted $m/z$,
\begin{equation}
  I_{\mathrm{merged}} = \sum_{\ell} I_{\ell},
  \qquad
  (m/z)_{\mathrm{merged}}
    = \frac{\sum_{\ell} I_{\ell}\,(m/z)_{\ell}}{\sum_{\ell} I_{\ell}},
  \label{eq:merge-rule}
\end{equation}
falling back to the arithmetic mean of $m/z$ when the total intensity is zero.
Peaks separated by more than $5$~ppm remain distinct, so a mixed view contains
at most $2 n_A$ peaks before merging, and no intensity renormalization is
applied afterwards. Competing spectra are drawn uniformly from the batch,
excluding the anchor itself and any spectrum whose neutral precursor mass lies
within $10$~ppm of the anchor's. The implementation additionally excludes
same-charge spectra whose precursor $m/z$ lies within $10$~ppm; because
neutral mass is determined by precursor $m/z$ and charge, this second criterion
excludes almost nothing beyond the neutral-mass one at equal charge, and is
retained as a conservative guard on the precursor representation the model
actually receives rather than as an independent chemical constraint. A fresh
competing spectrum and peak subset are drawn for each mixture view. If no
eligible spectrum is available the view remains anchor-only, which does not
occur in practice at the batch sizes used here.

\subsection{Self-supervised objective and optimization}
\label{app:ssl_objective}
\label{app:optimization}

\subsubsection{Prototype assignments and self-distillation targets}

The pretraining projection head maps a pooled representation to prototype
logits through
\begin{equation}
  \mathrm{Linear}(1024, 2048) \rightarrow \mathrm{GELU} \rightarrow
  \mathrm{Linear}(2048, 256) \rightarrow \ell_2 \rightarrow
  \mathrm{WN}\text{-}\mathrm{Linear}(256, 16{,}384; \text{no bias}),
  \label{eq:dino-head}
\end{equation}
where $\ell_2$ denotes $\ell_2$ normalization of the bottleneck feature and
$\mathrm{WN}$ denotes weight normalization. The final layer maps the
256-dimensional $\ell_2$-normalized bottleneck to $K = 16{,}384$ outputs, and
because it carries no bias its $K$ learned output weight vectors are the
prototypes of SwAV and DINOv2 \citep{caron2020swav,oquab2024dinov2}: output
$k$ is the inner product of the bottleneck feature with prototype $k$. The head's final weight scale
remains trainable, and the head is discarded for all downstream evaluation. Writing $g_\theta$ and $g_{\bar\theta}$ for the
student and teacher head applied to a pooled representation, student
assignments are temperature-scaled softmax distributions,
\begin{equation}
  p = \softmax\bigl(g_\theta(z) / \tau_s\bigr),
  \label{eq:student-assignment}
\end{equation}
while teacher assignments are balanced across the batch by Sinkhorn--Knopp
normalization of $\exp\bigl(g_{\bar\theta}(z)/\tau_t\bigr)$ with three
iterations, following DINOv2 \citep{oquab2024dinov2},
\begin{equation}
  q = \operatorname{SK}\bigl(g_{\bar\theta}(z) / \tau_t\bigr).
  \label{eq:teacher-assignment}
\end{equation}
Teacher assignments are detached, so no gradient flows through the teacher
branch, and after each optimizer step the teacher weights are updated as
$\bar\theta \leftarrow m\,\bar\theta + (1-m)\,\theta$. The teacher temperature
$\tau_t$ warms linearly from $0.03$ to $0.07$ over the first ten epochs and is
then held constant, the student temperature is $\tau_s = 0.1$, and the
momentum $m$ is cosine-scheduled from $0.998$ to $1.0$ over training. The
selected model sets both the masked-token (iBOT) and KoLeo loss weights to
zero and applies no token masking, so Equation~\ref{eq:loss-dion} is the
complete pretraining objective, and collapse is avoided by the
Sinkhorn--Knopp balancing and the teacher sharpening alone.

\subsubsection{Optimization}

The learning rate is warmed up linearly over 40,000 optimizer steps and then
cosine-decayed over a further 450,000 steps, with annealing applied per
optimizer step. Batch-size-dependent learning-rate scaling is disabled, so the
base learning rate is used directly. Weight decay is constant. Pretraining
runs on 64 GH200 96\,GB accelerators, 4 per node across 16 nodes, with a
per-device batch of 128 and no gradient accumulation, giving an effective
global batch of 8,192 spectra.
Table~\ref{tab:ssl-optimization} lists the settings.

\begin{table}[!htb]
\centering
\small
\caption{Self-supervised pretraining optimization settings.}
\label{tab:ssl-optimization}
\begin{tabular}{@{}ll@{}}
\toprule
Setting & Value \\
\midrule
Base learning rate & $8.0\times10^{-5}$ \\
Learning-rate scaling & Disabled \\
Warmup & 40,000 optimizer steps (linear, from 0) \\
Decay & Cosine over 450,000 steps to $1.0\times10^{-6}$ \\
Annealing granularity & Per optimizer step \\
Weight decay & 0.04 (constant) \\
Batch size per device & 128 \\
Devices & 16 nodes $\times$ 4 GH200 96\,GB (64 total) \\
Gradient accumulation & None \\
Effective global batch & 8,192 \\
Numerical precision & bf16 mixed \\
Training horizon & 300 epochs nominal, stopped after 284 \\
Checkpoint cadence & Every 10 epochs \\
Teacher EMA momentum & Cosine, $0.998 \rightarrow 1.0$ \\
Teacher temperature $\tau_t$ & $0.03 \rightarrow 0.07$ over 10 epochs, then constant \\
Student temperature $\tau_s$ & 0.1 \\
Teacher centering & Sinkhorn--Knopp, 3 iterations \\
\bottomrule
\end{tabular}
\end{table}

\label{app:refinement_config}

\subsection{Compute and training efficiency}
\label{app:compute}

The canonical pretraining run used 64 GH200 96\,GB accelerators, 4 per node
across 16 nodes, in bf16 mixed precision with an effective global batch of
8,192 spectra (Table~\ref{tab:ssl-optimization}). Allocated device memory was
stable at roughly $90\%$ of the $96$\,GB available on each accelerator for the
duration of training. Completing the 284 trained
epochs took about 75 hours of wall-clock time, approximately 4,800
accelerator-hours. These figures
describe the canonical run alone. The ablations run shorter horizons under
separate configurations.

\subsection{Model selection}
\label{app:model_selection}

The released pretraining checkpoint, epoch 284 of the canonical run, was
selected manually by inspecting the online validation probes of the objective ablations: strict
pair discrimination and broad retrieval (Figure~\ref{fig:objective-ablation})
and the end-AA probes (Appendix~\ref{app:objective_ablations}). The teacher temperature and
momentum schedules were chosen with the collapse diagnostics of
Appendix~\ref{app:collapse}. Main-text similarity results use cosine
similarity of the global representation. For the mechanistic probes,
Appendix~\ref{app:dual_view_interventions} also reports Euclidean distance and
the Jensen--Shannon divergence of the projection-head outputs. De novo checkpoints are selected by
maximum peptide precision on the validation split of their training corpus:
\DSMSKBVal{} spectra for \DatasetShort{mskbfinal} and \DSKitchenSinkVal{} for
\DatasetShort{kitchensinkvfive}. De novo training and decoding
hyperparameters, including greedy decoding, the 200-peak cap and the 50~ppm
precursor tolerance, were fixed in advance and not tuned in this work
(Appendix~\ref{app:denovo_decoding}). Test splits were used only for final
evaluation.

\label{app:frozen_configs}

%% file: appendix/C_embed_eval.tex
\section{Embedding evaluation and additional analyses}
\label{app:embedding_analysis}

\subsection{Objective ablations}
\label{app:objective_ablations}

The end-AA probe is a small binary diagnostic tracked during development:
tryptic peptides end in K or R, and the probe predicts which from the
representation, by nearest neighbours or a linear classifier.

\newcommand{\abllegendapp}[2]{%
  \textcolor{#1}{\rule[0.28ex]{1.1em}{1.1pt}}\,\textcolor{#1}{#2}\;}
\begin{figure}[!htb]
\centering
{\footnotesize
 \abllegendapp{ablPure}{$\blacksquare$}Pure mixture \qquad
 \abllegendapp{ablFree}{$\blacktriangle$}Mixture-free \qquad
 \abllegendapp{ablDual}{$\bullet$}Dual objective\par}
\vspace{0.4em}
\begin{subfigure}[t]{0.40\linewidth}
  \centering
  \includegraphics[width=\linewidth,trim={0 0 0 23pt},clip]{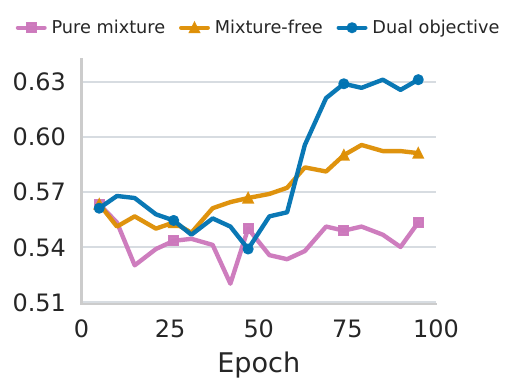}
  \caption{End-AA kNN}
\end{subfigure}\hspace{0.03\linewidth}
\begin{subfigure}[t]{0.40\linewidth}
  \centering
  \includegraphics[width=\linewidth,trim={0 0 0 23pt},clip]{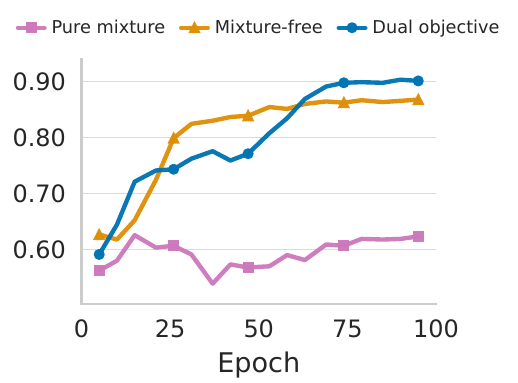}
  \caption{End-AA linear probe}
\end{subfigure}
\caption{End-AA probes during pretraining of the 100-epoch ablation
checkpoints, complementing Figure~\ref{fig:objective-ablation}. The dual
objective leads both probes in the second half of pretraining.}
\label{fig:objective-ablation-endaa}
\end{figure}

The three runs are matched approximately 100-epoch runs on the same bacterial
corpus, architecture, optimizer, schedule, preprocessing and 200-peak cap,
evaluated with the same compact validation probe suite: strict pair AUROC and
broad retrieval mAP (Section~\ref{sec:geometry},
Figure~\ref{fig:objective-ablation}) and the end-AA probes
(Figure~\ref{fig:objective-ablation-endaa}).
The probes read out the pooled representation of the EMA teacher on fixed, compact full-charge subsets rather than the full validation sets. The canonical run
evaluates them every 12,000 optimizer steps. The ablation runs log them at
differing cadences, so their curves differ in sampling density, but all are
aligned by training epoch and span the same 100 epochs. Strict pair AUROC and broad retrieval mAP are each
averaged over subsets of three validation sets: \DatasetShort{ninespecies}
validation, the \DatasetShort{bacterialssl} development subset, and
\DatasetShort{kingdoms} validation. The end-AA probes use a dedicated set of
4,200 training, 900 validation and 900 test peptides. The kNN probe ($k=64$,
cosine similarity) is scored on its validation rows, and the linear probe is
fitted on its training rows and scored on its test rows. The three objectives differ only in
how precursor conditioning and mixture peaks are allocated across view scales.
The pure mixture objective gives the precursor-conditioned global student
views mixed fragment peaks and has no precursor-withheld partial-view objective. The
mixture-free precursor-inference control introduces no distractor peaks at
all: clean $95\%$ global student views receive the null precursor and predict
clean precursor-conditioned teacher targets, while clean $60\%$ partial student
views retain the true precursor, with the two losses weighted equally. The
dual objective is the method of Section~\ref{sec:method}.

For de novo transfer (Table~\ref{tab:denovo-objective-ablation}), each
pretrained encoder is paired with a fresh 94.6M-parameter decoder and trained
for 80 epochs on \DatasetShort{mskbfinal} training data at a 200-peak cap with
precursor conditioning, either with the encoder frozen or end to end. The
frozen control is a randomly initialized encoder, and the end-to-end control is
trained from scratch. Each run reports its maximum validation peptide
precision over the 80 epochs. The distributed validation sampler evaluates
10,016 items by repeating 12 of the 10,004 validation spectra, identically for
every run, so the comparison is unaffected.

With the encoder frozen, the dual objective gives the highest peptide precision, followed by the pure mixture objective, the mixture-free control and a randomly initialized encoder (Table~\ref{tab:denovo-objective-ablation}). End-to-end fine-tuning preserves this ordering, and every objective improves on training from scratch. Because fine-tuning adapts the encoder under full peptide supervision, it compresses the differences between objectives. The representations themselves differ substantially, as the frozen column and the probes of Figure~\ref{fig:objective-ablation} show, and the dual objective leads on every fragment-sensitive measure.

\input{code/generated/tables/validation_denovo_objective_ablation}

Representation geometry is additionally summarized by unsupervised embedding
quality evaluation \citep{pmlr-v221-tsitsulin23a}, a family of label-free
diagnostics that includes rank-based measures such as RankMe
\citep{pmlr-v202-garrido23a}, eigenspectrum-decay measures such as
$\alpha$-ReQ \citep{agrawal2022alphareq}, and NESum
\citep{pmlr-v162-he22c}. We report normalized RankMe and self-cluster.

\begin{figure}[!htb]
\centering
\begin{minipage}[t]{0.48\linewidth}
\centering
\includegraphics[width=\linewidth]{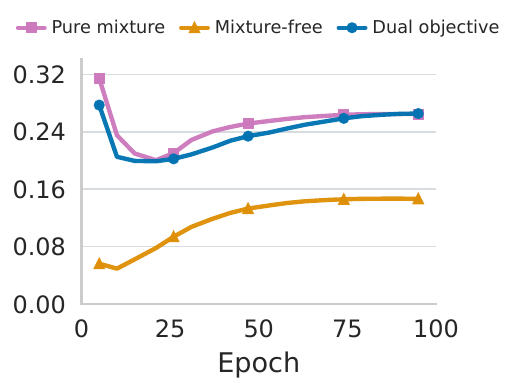}
\end{minipage}\hfill
\begin{minipage}[t]{0.48\linewidth}
\centering
\includegraphics[width=\linewidth]{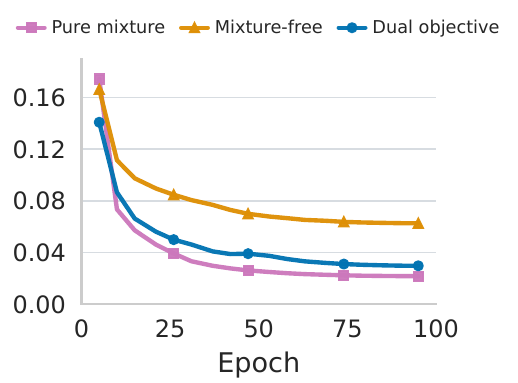}
\end{minipage}
\caption{Unsupervised embedding quality evaluation for the objective ablation:
normalized RankMe \citep{pmlr-v202-garrido23a} and self-cluster
\citep{pmlr-v221-tsitsulin23a}. These are coarse checks for pathological
geometry and are not monotonic measures of representation quality, so they are
not used to rank the objectives. In
particular, the pure mixture objective shows no dimensional collapse: its effective
rank is not low. Its failure is representational collapse onto precursor
information, which these diagnostics cannot detect.}
\label{fig:objective-ablation-health}
\end{figure}

\subsection{Mechanistic validation of the dual latent-prediction views}
\label{app:dual_view_interventions}

Both probes evaluate the same frozen checkpoint on the
\DatasetFull{ninespecies} validation retrieval artifact described in
Appendix~\ref{app:datasets}: \DSNineSpeciesVal{} spectra forming 900
repeated-peptide groups, 100 groups from each of the nine species, with at
most ten spectra per group. Peptide backbones occurring in the reserved
NineSpecies test benchmark were excluded under uppercase amino-acid
normalization, so the probe population is disjoint from that benchmark.
Spectra use the standard preprocessing of
Appendix~\ref{app:model_training_config} with the peak cap set to 200.

Pairing is deterministic. Up to 500 anchors are drawn per species, giving
4,488 directed $(A,B)$ pairs. A candidate $B$ must carry a different peptide
identity from $A$, be a different spectrum, differ by more than $10$~ppm in
neutral precursor mass, and, when the charges match, differ by more than
$10$~ppm in precursor $m/z$. Anchors are stratified by species but $B$ is
drawn from the full artifact and need not share the anchor's species. All
reported values are unweighted means over the nine anchor-species partitions.

For the precursor-swap probe the mixture is formed once per pair and embedded
twice, so only the precursor differs between the two embeddings. The competing
spectrum contributes at most as many peaks as the anchor retains
(Equation~\ref{eq:competing-size}), matching the size asymmetry of the
training-time construction; peaks within $5$~ppm are merged with summed
intensity and intensity-weighted $m/z$, and no intensity renormalization
follows. This bound, together with the one-sided objective in which the clean
anchor is always the target, is the reason the competing-precursor condition
is the harder of the two: the competing source is systematically the weaker
contributor to the mixture and is never itself a prediction target. Both are
deliberate properties of a representation-learning objective. Removing the
size bound and adding a symmetric term, so that either source must be
recoverable under its own precursor, is the natural extension toward a
decomposition model, and is not attempted here. For the partial-view probe, two $60\%$ peak subsets of
the anchor are sampled independently and embedded under the null precursor
state, with no competing peaks mixed in.

Tables~\ref{tab:validation-mechanism-mixed-source}
and~\ref{tab:validation-mechanism-null-local} report every readout. The
margin is the mean gap between the correct and incorrect source in the units
of each readout, so it is comparable within a row but not across rows.

\input{code/generated/tables/validation_mechanism_mixed_source}
\input{code/generated/tables/validation_mechanism_null_local}

The two probes above use constructed mixtures. We also test
precursor-conditioned selection on measured co-isolated spectra, in which the
instrument isolated two peptides together
\citep{houel2010chimera,michalski2011detectable}, and replace the two-point
swap with a sweep of the conditioning mass.

A spectrum qualifies when two peptides were separately identified in the same
run with the same precursor charge, eluted within 60~s of each other, and the
second precursor lies inside the recorded isolation window of the first,
targeted precursor. Each peptide must be identified in at least two other
scans, which give its reference representation, embedded under its own
precursor. We hold the peaks fixed, sweep the conditioning mass in 0.005~Da
steps, and record the cosine margin: the cosine similarity to the reference of
the targeted peptide~A minus that to the co-isolated peptide~B.

\begin{figure}[!htb]
\centering
\includegraphics[width=.5\linewidth]{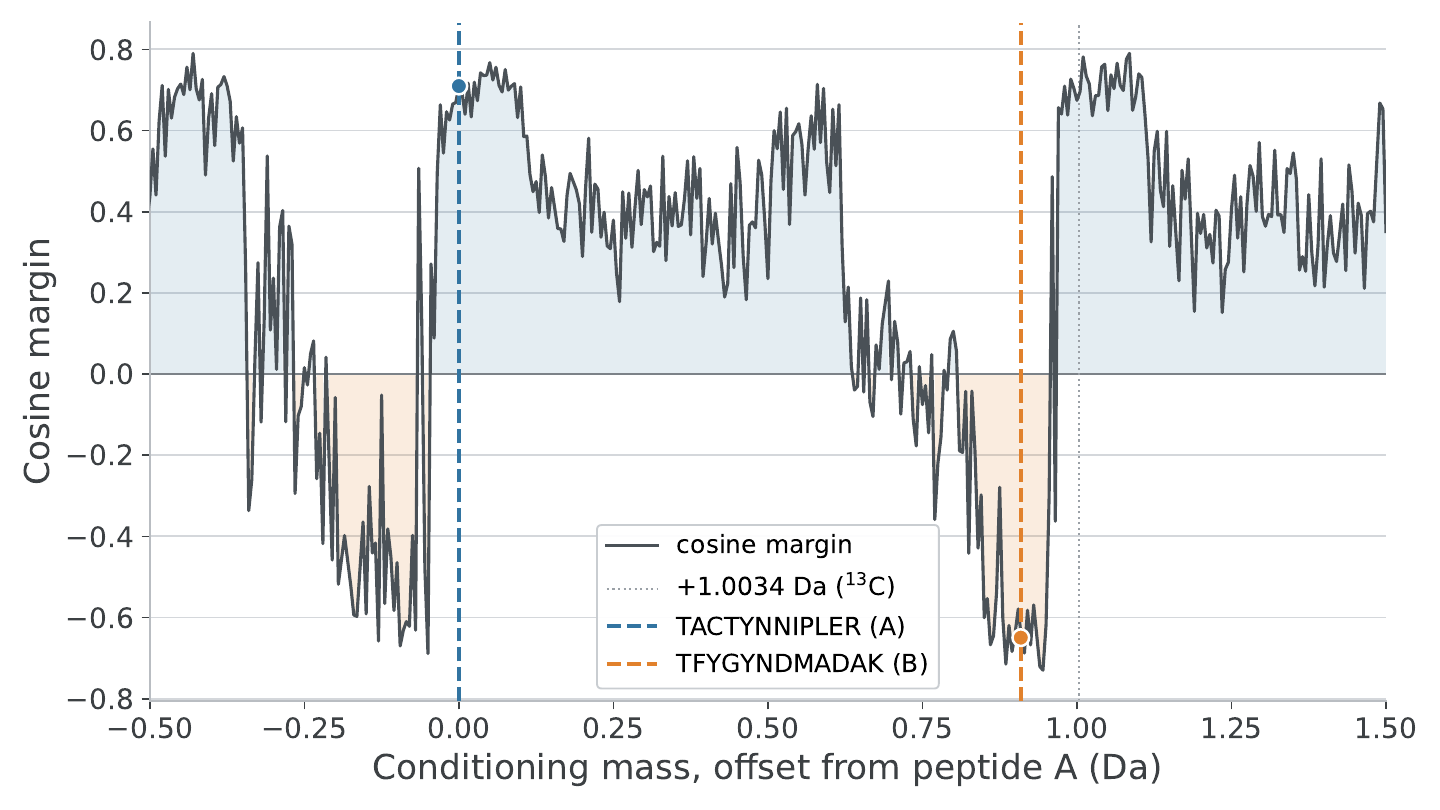}
\caption{Conditioning-mass sweep on one measured spectrum containing the
co-isolated peptides TACTYNNIPLER (A, targeted) and TFYGYNDMADAK (B). The
peaks are fixed and only the conditioning mass varies. A positive margin means
the representation is closer to A's reference than to B's. Dashed lines mark
the true masses of A and B, and the dotted line the first $^{13}$C isotope
of A.}
\label{fig:conditioning-mass-switch}
\end{figure}

In the spectrum of Figure~\ref{fig:conditioning-mass-switch}, conditioning on
each true precursor mass selects the corresponding peptide, and the two
references are nearly orthogonal (cosine 0.009). Selection also repeats at the
$^{13}$C spacing: the margin favours A again near A~$+$~1.003~Da and B near
B~$-$~1.003~Da, consistent with the one-isotope errors common in precursor
assignment. Across 45 co-isolated pairs from one run, the margin changes sign
between the two true masses in 6, three of which rest on a single reference
scan for one peptide, and in no pair closer than 0.2~Da.
In the other pairs,
conditioning on B moves the representation toward B without making it closer
to B than to A. B is the harder direction here for the same reason as in the
constructed probes (Table~\ref{tab:validation-mechanism-mixed-source}): it is
the weaker contributor to the spectrum, and training never makes the competing
source a target.

\subsection{Complete representation-geometry diagnostics}
\label{app:complete_geometry_tables}

The main tables emphasize precursor-controlled retrieval AP and static-pair
AUROC. For completeness, the tables below additionally report unfiltered
global retrieval and static-pair AP from the same evaluation runs. Global retrieval admits precursor-mass and charge shortcuts, so we
treat it as a diagnostic rather than a headline comparison.

\input{code/generated/tables/validation_representation_all}

\input{code/generated/tables/validation_representation_bacterial_complete}
\input{code/generated/tables/validation_representation_ninespecies_complete}
\input{code/generated/tables/validation_representation_kingdoms_complete}
\input{code/generated/tables/validation_gleams_bacterial_complete}
\input{code/generated/tables/validation_gleams_ninespecies_complete}
\input{code/generated/tables/validation_gleams_kingdoms_complete}

%% file: code/generated/tables/validation_denovo_objective_ablation.tex
\begin{table}[!htb]
\centering
\small
\caption{De novo transfer from the 100-epoch ablation checkpoints on the \DatasetShort{mskbfinal} development-\emph{validation} split. The last row is the control without pretraining: a randomly initialized frozen encoder, and a model trained from scratch.}
\label{tab:denovo-objective-ablation}
\begin{threeparttable}
\begin{tabular}{@{}lS[table-format=1.3]S[table-format=1.3]@{}}
\toprule
 & \multicolumn{2}{c}{Peptide precision} \\
\cmidrule(lr){2-3}
{Pretraining objective} & {Frozen encoder} & {End-to-end} \\
\midrule
Dual objective & \textbf{0.675} & \textbf{0.782} \\
Pure mixture & 0.638 & 0.778 \\
Mixture-free control & 0.610 & 0.764 \\
No pretraining & 0.582 & 0.744 \\
\bottomrule
\end{tabular}
\end{threeparttable}
\end{table}

%% file: code/generated/tables/validation_mechanism_mixed_source.tex
\begin{table}[!htb]
\centering
\caption{Precursor-conditioned source selection by \modelname{} on the NineSpecies validation set. The mixture $A \cup B$ of an anchor $A$ and a competing spectrum $B$ is held fixed and only the precursor changes. A trial is correct when the representation lies closer to the clean representation of the conditioned source than to that of the other source.}
\label{tab:validation-mechanism-mixed-source}
\begin{threeparttable}
\begin{tabular}{@{}lS[table-format=1.3]S[table-format=1.3]S[table-format=1.3]S[table-format=2.3]@{}}
\toprule
 & \multicolumn{3}{c}{Correct-source accuracy} & \multicolumn{1}{c}{Separation} \\
\cmidrule(lr){2-4}
\cmidrule(lr){5-5}
{Readout} & {Precursor of $A$} & {Precursor of $B$} & {Both} & {Margin} \\
\midrule
Backbone cosine (selected) & 0.997 & 0.929 & 0.926 & 0.767 \\
Backbone Euclidean & 0.999 & 0.847 & 0.846 & 18.860 \\
Projection head, Jensen--Shannon & 0.998 & 0.869 & 0.867 & 0.004 \\
\bottomrule
\end{tabular}
\begin{tablenotes}[flushleft]
\footnotesize
\item[] \textit{The margin is expressed in the units of each readout: cosine similarity, Euclidean distance, and Jensen--Shannon divergence.}
\end{tablenotes}
\end{threeparttable}
\end{table}

%% file: code/generated/tables/validation_mechanism_null_local.tex
\begin{table}[!htb]
\centering
\caption{Source recovery by \modelname{} from independent precursor-withheld partial views, NineSpecies validation set. Two independently sampled 60\% peak subsets of each anchor are embedded under the null precursor state. A view is correct when it lies closer to the clean representation of its own anchor $A$ than to that of a different-peptide spectrum $B$.}
\label{tab:validation-mechanism-null-local}
\begin{threeparttable}
\begin{tabular}{@{}lS[table-format=1.3]S[table-format=1.3]S[table-format=1.3]S[table-format=1.3]S[table-format=1.3]@{}}
\toprule
 & \multicolumn{4}{c}{Correct-source accuracy} & \multicolumn{1}{c}{Separation} \\
\cmidrule(lr){2-5}
\cmidrule(lr){6-6}
{Readout} & {View 1} & {View 2} & {Both} & {Mean} & {Margin} \\
\midrule
Backbone cosine (selected) & 0.987 & 0.984 & 0.978 & 0.986 & 0.498 \\
\bottomrule
\end{tabular}
\end{threeparttable}
\end{table}

%% file: code/generated/tables/validation_representation_all.tex
\begin{table*}[!htb]
\centering
\small
\caption{Native representation geometry across test benchmarks over all precursor charges. Unlike the charge-2--4 subsets of Table~\ref{tab:gleams-matched-all}, these populations retain slightly more spectra and exclude GLEAMS, which embeds only charges 2--4. All metrics compare only spectra with the same precursor charge and precursor mass within 10 ppm.}
\label{tab:native-representation-all}
\begin{threeparttable}
\begin{tabular}{@{}lS[table-format=1.3]S[table-format=1.3]S[table-format=1.3]S[table-format=1.3]S[table-format=1.3]S[table-format=1.3]@{}}
\toprule
 & \multicolumn{2}{c}{Bacterial} & \multicolumn{2}{c}{NineSpecies} & \multicolumn{2}{c}{Kingdoms} \\
\cmidrule(lr){2-3}
\cmidrule(lr){4-5}
\cmidrule(lr){6-7}
 & \multicolumn{1}{c}{Retrieval} & \multicolumn{1}{c}{Pairs} & \multicolumn{1}{c}{Retrieval} & \multicolumn{1}{c}{Pairs} & \multicolumn{1}{c}{Retrieval} & \multicolumn{1}{c}{Pairs} \\
{Method} & {mAP} & {AUROC} & {mAP} & {AUROC} & {mAP} & {AUROC} \\
\midrule
Binned spectral angle & 0.913 & 0.941 & 0.972 & 0.954 & 0.908 & 0.871 \\
Random Transformer & 0.714 & 0.791 & 0.823 & 0.704 & 0.710 & 0.637 \\
\addlinespace
\multicolumn{7}{@{}l}{\itshape Peptide-supervised} \\
Casanovo 4.0 & 0.880 & 0.896 & 0.933 & 0.858 & 0.838 & 0.746 \\
\addlinespace
\multicolumn{7}{@{}l}{\itshape Self-supervised} \\
InstaNovo-FM & 0.847 & 0.862 & 0.877 & 0.762 & 0.806 & 0.708 \\
\modelname{} & 0.893 & 0.927 & 0.922 & 0.864 & 0.824 & 0.748 \\
\addlinespace[0.6em]
\multicolumn{7}{@{}l}{\itshape Peptide-supervised contrastive} \\
Scratch init. (1\%) & 0.907 & 0.932 & 0.970 & 0.946 & 0.909 & 0.873 \\
\modelname{} init. (1\%) & 0.920 & 0.952 & 0.971 & 0.952 & 0.914 & 0.887 \\
Scratch init. (10\%) & 0.926 & 0.958 & \textbf{0.977} & 0.956 & 0.933 & 0.901 \\
\modelname{} init. (10\%) & \textbf{0.930} & \textbf{0.966} & \textbf{0.977} & \textbf{0.965} & \textbf{0.942} & \textbf{0.926} \\
\bottomrule
\end{tabular}
\end{threeparttable}
\end{table*}

%% file: code/generated/tables/validation_representation_bacterial_complete.tex
\begin{table}[!htb]
\centering
\small
\caption{Complete native-geometry diagnostics on the bacterial test set.}
\label{tab:native-bacterial-complete}
\begin{threeparttable}
\begin{tabular}{@{}lS[table-format=1.3]S[table-format=1.3]S[table-format=1.3]S[table-format=1.3]@{}}
\toprule
 & \multicolumn{2}{c}{Retrieval} & \multicolumn{2}{c}{Static pairs} \\
\cmidrule(lr){2-3}
\cmidrule(lr){4-5}
{Method} & {Broad mAP} & {AP} & {AP} & {AUROC} \\
\midrule
Binned spectral angle & 0.527 & 0.913 & 0.934 & 0.941 \\
Random Transformer & 0.126 & 0.714 & 0.804 & 0.791 \\
Casanovo 4.0 & 0.416 & 0.880 & 0.900 & 0.896 \\
InstaNovo-FM & 0.377 & 0.847 & 0.876 & 0.862 \\
\modelname{} & 0.492 & 0.893 & 0.926 & 0.927 \\
\addlinespace
\multicolumn{5}{@{}l}{\itshape Supervised metric learning} \\
Scratch 10\% & 0.544 & 0.926 & 0.947 & 0.958 \\
\modelname{} 10\% f.t. & 0.577 & 0.930 & 0.960 & 0.966 \\
\bottomrule
\end{tabular}
\end{threeparttable}
\end{table}

%% file: code/generated/tables/validation_representation_ninespecies_complete.tex
\begin{table}[!htb]
\centering
\small
\caption{Complete native-geometry diagnostics on the NineSpecies V2 test set.}
\label{tab:native-ninespecies-complete}
\begin{threeparttable}
\begin{tabular}{@{}lS[table-format=1.3]S[table-format=1.3]S[table-format=1.3]S[table-format=1.3]@{}}
\toprule
 & \multicolumn{2}{c}{Retrieval} & \multicolumn{2}{c}{Static pairs} \\
\cmidrule(lr){2-3}
\cmidrule(lr){4-5}
{Method} & {Broad mAP} & {AP} & {AP} & {AUROC} \\
\midrule
Binned spectral angle & 0.784 & 0.972 & 0.939 & 0.954 \\
Random Transformer & 0.157 & 0.823 & 0.726 & 0.704 \\
Casanovo 4.0 & 0.524 & 0.933 & 0.862 & 0.858 \\
InstaNovo-FM & 0.417 & 0.877 & 0.784 & 0.762 \\
\modelname{} & 0.611 & 0.922 & 0.868 & 0.864 \\
\addlinespace
\multicolumn{5}{@{}l}{\itshape Supervised metric learning} \\
Scratch 10\% & 0.796 & 0.977 & 0.937 & 0.956 \\
\modelname{} 10\% f.t. & 0.807 & 0.977 & 0.952 & 0.965 \\
\bottomrule
\end{tabular}
\end{threeparttable}
\end{table}

%% file: code/generated/tables/validation_representation_kingdoms_complete.tex
\begin{table}[!htb]
\centering
\small
\caption{Complete native-geometry diagnostics on the Kingdoms test set.}
\label{tab:native-kingdoms-complete}
\begin{threeparttable}
\begin{tabular}{@{}lS[table-format=1.3]S[table-format=1.3]S[table-format=1.3]S[table-format=1.3]@{}}
\toprule
 & \multicolumn{2}{c}{Retrieval} & \multicolumn{2}{c}{Static pairs} \\
\cmidrule(lr){2-3}
\cmidrule(lr){4-5}
{Method} & {Broad mAP} & {AP} & {AP} & {AUROC} \\
\midrule
Binned spectral angle & 0.499 & 0.908 & 0.879 & 0.871 \\
Random Transformer & 0.083 & 0.710 & 0.661 & 0.637 \\
Casanovo 4.0 & 0.324 & 0.838 & 0.780 & 0.746 \\
InstaNovo-FM & 0.323 & 0.806 & 0.753 & 0.708 \\
\modelname{} & 0.442 & 0.824 & 0.767 & 0.748 \\
\addlinespace
\multicolumn{5}{@{}l}{\itshape Supervised metric learning} \\
Scratch 10\% & 0.600 & 0.933 & 0.913 & 0.901 \\
\modelname{} 10\% f.t. & 0.655 & 0.942 & 0.934 & 0.926 \\
\bottomrule
\end{tabular}
\end{threeparttable}
\end{table}

%% file: code/generated/tables/validation_gleams_bacterial_complete.tex
\begin{table}[!htb]
\centering
\small
\caption{Complete charge-2--4 matched diagnostics on the bacterial test set.}
\label{tab:gleams-bacterial-complete}
\begin{threeparttable}
\begin{tabular}{@{}lS[table-format=1.3]S[table-format=1.3]S[table-format=1.3]S[table-format=1.3]@{}}
\toprule
 & \multicolumn{2}{c}{Retrieval} & \multicolumn{2}{c}{Static pairs} \\
\cmidrule(lr){2-3}
\cmidrule(lr){4-5}
{Method} & {Broad mAP} & {AP} & {AP} & {AUROC} \\
\midrule
Binned spectral angle & 0.537 & 0.915 & 0.938 & 0.943 \\
Casanovo 4.0 & 0.424 & 0.881 & 0.903 & 0.898 \\
InstaNovo-FM & 0.384 & 0.847 & 0.880 & 0.864 \\
GLEAMS & 0.475 & 0.916 & 0.946 & 0.948 \\
\modelname{} & 0.506 & 0.895 & 0.929 & 0.928 \\
\addlinespace
\multicolumn{5}{@{}l}{\itshape Supervised metric learning} \\
Scratch 10\% & 0.559 & 0.928 & 0.950 & 0.960 \\
\modelname{} 10\% f.t. & 0.589 & 0.933 & 0.962 & 0.968 \\
\bottomrule
\end{tabular}
\end{threeparttable}
\end{table}

%% file: code/generated/tables/validation_gleams_ninespecies_complete.tex
\begin{table}[!htb]
\centering
\small
\caption{Complete charge-2--4 matched diagnostics on the NineSpecies V2 test set.}
\label{tab:gleams-ninespecies-complete}
\begin{threeparttable}
\begin{tabular}{@{}lS[table-format=1.3]S[table-format=1.3]S[table-format=1.3]S[table-format=1.3]@{}}
\toprule
 & \multicolumn{2}{c}{Retrieval} & \multicolumn{2}{c}{Static pairs} \\
\cmidrule(lr){2-3}
\cmidrule(lr){4-5}
{Method} & {Broad mAP} & {AP} & {AP} & {AUROC} \\
\midrule
Binned spectral angle & 0.786 & 0.975 & 0.939 & 0.954 \\
Casanovo 4.0 & 0.526 & 0.932 & 0.863 & 0.859 \\
InstaNovo-FM & 0.419 & 0.874 & 0.785 & 0.763 \\
GLEAMS & 0.746 & 0.976 & 0.951 & 0.962 \\
\modelname{} & 0.612 & 0.926 & 0.869 & 0.865 \\
\addlinespace
\multicolumn{5}{@{}l}{\itshape Supervised metric learning} \\
Scratch 10\% & 0.797 & 0.978 & 0.938 & 0.957 \\
\modelname{} 10\% f.t. & 0.808 & 0.978 & 0.953 & 0.965 \\
\bottomrule
\end{tabular}
\end{threeparttable}
\end{table}

%% file: code/generated/tables/validation_gleams_kingdoms_complete.tex
\begin{table}[!htb]
\centering
\small
\caption{Complete charge-2--4 matched diagnostics on the Kingdoms test set.}
\label{tab:gleams-kingdoms-complete}
\begin{threeparttable}
\begin{tabular}{@{}lS[table-format=1.3]S[table-format=1.3]S[table-format=1.3]S[table-format=1.3]@{}}
\toprule
 & \multicolumn{2}{c}{Retrieval} & \multicolumn{2}{c}{Static pairs} \\
\cmidrule(lr){2-3}
\cmidrule(lr){4-5}
{Method} & {Broad mAP} & {AP} & {AP} & {AUROC} \\
\midrule
Binned spectral angle & 0.503 & 0.905 & 0.879 & 0.871 \\
Casanovo 4.0 & 0.327 & 0.838 & 0.780 & 0.746 \\
InstaNovo-FM & 0.326 & 0.805 & 0.753 & 0.708 \\
GLEAMS & 0.440 & 0.903 & 0.895 & 0.876 \\
\modelname{} & 0.445 & 0.821 & 0.767 & 0.748 \\
\addlinespace
\multicolumn{5}{@{}l}{\itshape Supervised metric learning} \\
Scratch 10\% & 0.605 & 0.930 & 0.914 & 0.902 \\
\modelname{} 10\% f.t. & 0.659 & 0.940 & 0.934 & 0.926 \\
\bottomrule
\end{tabular}
\end{threeparttable}
\end{table}

%% file: appendix/D_downstream_transfer.tex
\section{Downstream transfer setup}
\label{app:downstream_transfer}

\subsection{Auxiliary spectrum-level task definitions}
\label{app:downstream_tasks}

Spectral quality assessment, chimericity, and oxidized-methionine prediction
share a single binary prediction head applied to the pooled global
representation $z \in \mathbb{R}^{1024}$,
\begin{equation}
\begin{aligned}
  z \;\rightarrow\;
  &\mathrm{Linear}(1024, 128) \rightarrow \mathrm{GELU}
   \rightarrow \mathrm{LayerNorm}(128) \\
  \rightarrow\;
  &\mathrm{Linear}(128, 64) \rightarrow \mathrm{GELU}
   \rightarrow \mathrm{LayerNorm}(64)
   \rightarrow \mathrm{Linear}(64, 1),
\end{aligned}
\label{eq:binary-head}
\end{equation}
whose scalar output is a binary logit. Dropout is disabled throughout this
head.

\input{code/generated/tables/test_auxiliary_auc}

\subsection{Supervised peptide metric learning}
\label{app:metric_learning}

A freshly initialized projection head is applied to the pooled global
representation, upstream of the discarded pretraining prototype head,
\begin{equation}
  z \;\rightarrow\;
  \mathrm{Linear}(1024, 1024) \rightarrow \mathrm{GELU}
  \rightarrow \mathrm{Linear}(1024, 128)
  \rightarrow \ell_2\text{-normalization},
  \label{eq:metric-head}
\end{equation}
producing unit-norm 128-dimensional embeddings.

Writing $v_i$ for the normalized embedding of example $i$, $y_i$ for its
peptide label, and $P(i) = \{p \neq i : y_p = y_i\}$ for its set of
in-batch positives, the supervised contrastive objective
\citep{khosla2020supcon} is
\begin{equation}
  \mathcal{L}_i = -\frac{1}{|P(i)|} \sum_{p \in P(i)}
    \log \frac{\exp\bigl(v_i^{\top} v_p / \tau\bigr)}
              {\sum_{a \neq i} \exp\bigl(v_i^{\top} v_a / \tau\bigr)},
  \qquad \tau = 0.07,
  \label{eq:supcon}
\end{equation}
averaged over anchors with at least one positive. Different-peptide examples
enter as negatives through the denominator, so no separate negative-loss term
is required.

%% file: code/generated/tables/test_auxiliary_auc.tex
\begin{table}[!htb]
\centering
\small
\caption{Auxiliary binary spectrum-level tasks on held-out test sets (AUROC). Frozen representations and end-to-end adaptation are reported as separate regimes.}
\label{tab:test-auxiliary-auc}
\begin{threeparttable}
\begin{tabular}{@{}lS[table-format=1.3]S[table-format=1.3]S[table-format=1.3]@{}}
\toprule
 & \multicolumn{3}{c}{Task AUROC} \\
\cmidrule(lr){2-4}
{Representation / initialization} & {SQA} & {Chimericity} & {Oxidized Met} \\
\midrule
\multicolumn{4}{@{}l}{\itshape Frozen transfer} \\
Binned + precursor metadata & 0.737 & 0.708 & 0.654 \\
Random Transformer & 0.753 & 0.760 & 0.642 \\
Casanovo 4.0 & 0.807 & 0.807 & 0.795 \\
InstaNovo-FM & 0.810 & 0.827 & 0.835 \\
\modelname{} (conditioned) & 0.805 & 0.704 & 0.730 \\
\addlinespace
\multicolumn{4}{@{}l}{\itshape Fine-tuned transfer} \\
Scratch, end-to-end & 0.783 & 0.762 & 0.749 \\
\modelname{}, end-to-end & 0.860 & 0.799 & 0.855 \\
\bottomrule
\end{tabular}
\end{threeparttable}
\end{table}

%% file: appendix/E_denovo.tex
\section{De novo sequencing and dense transfer}
\label{app:denovo_dense}

\subsection{De novo encoder--decoder architecture}
\label{app:denovo_architecture}

The peptide decoder is a freshly initialized Casanovo-style autoregressive
Transformer that performs causal self-attention over the peptide prefix and
cross-attention over the encoder peak-token memory $\{h_i\}$, whose dimension
matches the encoder model width. Target sequences use a 45-token
vocabulary: 43 residue and N-terminal-modification tokens, one padding token,
and an end-of-sequence token. A precursor token is prepended to the decoder
target stream, formed by summing a sinusoidal encoding of the neutral
precursor mass with a learned charge embedding over charges 1--10, and
sinusoidal positional encoding is applied over the precursor token together
with the peptide target sequence. Table~\ref{tab:denovo-decoder} lists the
configuration.

\begin{table}[!htb]
\centering
\small
\caption{De novo peptide decoder configuration.}
\label{tab:denovo-decoder}
\begin{tabular}{@{}ll@{}}
\toprule
Component & Setting \\
\midrule
Decoder blocks & 9 \\
Model dimension & 1024 \\
Attention heads & 8 \\
Feed-forward width & 1024 \\
Dropout & 0.25 \\
Encoder memory dimension & 1024 \\
Self-attention & Causal, over peptide tokens \\
Cross-attention & Encoder peak-token memory $\{h_i\}$ \\
Output projection & Linear(1024, 45) to token logits \\
\addlinespace
Target vocabulary & 45 tokens (43 residue/N-terminal modification, padding, \textsc{[eos]}) \\
Decoder precursor token & Sinusoidal precursor-mass encoding $+$ learned charge embedding (1--10) \\
Positional encoding & Sinusoidal, over precursor token and peptide sequence \\
\bottomrule
\end{tabular}
\end{table}

\subsection{Autoregressive training and fine-tuning}
\label{app:denovo_training}

Supervised de novo training minimizes teacher-forced categorical
cross-entropy over peptide tokens without label smoothing, using the Adam
optimizer with a cosine warmup and decay schedule.
Table~\ref{tab:denovo-optimization} lists the settings, which are held fixed
across scratch and \modelname{}-initialized conditions.

\begin{table}[!htb]
\centering
\small
\caption{De novo supervised training settings, held fixed across
initialization conditions.}
\label{tab:denovo-optimization}
\begin{tabular}{@{}ll@{}}
\toprule
Setting & Value \\
\midrule
Objective & Teacher-forced token cross-entropy \\
Label smoothing & None \\
Optimizer & Adam \\
Learning rate & $1\times10^{-4}$ \\
Weight decay & None \\
Gradient clipping & 2.0 \\
Schedule & Cosine warmup and decay \\
\bottomrule
\end{tabular}
\end{table}

\subsection{Unlabeled yeast evaluation}
\label{app:yeast_unlabeled}

The unlabeled benchmark uses one public yeast run from PRIDE accession
PXD026806, file \texttt{MDR27022021\_S01\_2\_DDA01}, converted to peak lists
with ThermoRawFileParser 1.4.5, giving 39,460 spectra. No ground-truth peptide
enters the pipeline at any stage. All three models receive the identical
39,460-scan roster under the same identifiers and decode greedily with a single
beam.

A prediction is scored by whether it occurs as a contiguous substring of the
yeast proteome (UniProt UP000002311, 6,067 sequences), concatenated with
separators so that no match may span two proteins. Modifications are stripped
before matching, and isoleucine and leucine are collapsed to a single symbol on
both sides, since the two are isobaric and therefore indistinguishable in
principle. Because short strings occur by chance in a reference of this size,
the same procedure is repeated against two decoy references built from the same
sequences, one with each protein reversed and one with each protein shuffled
while preserving its amino-acid composition. These bound the match rate
attainable without real signal.

\modelname{} and InstaNovo return a prediction for every spectrum. Casanovo
declines 3,934, of which 3,188 carry too few peaks and 746 exceed its supported
charge range. These are retained as explicit declines, so the match fraction is
reported both over emitted predictions and over all submitted spectra.
Table~\ref{tab:denovo-yeast-unlabeled-complete} also gives coverage, the
tryptic-restricted match fraction, the composition-preserving decoy, and the
fraction of calls carrying a modification.

\input{code/generated/tables/test_denovo_yeast_unlabeled_complete}

\subsection{Kingdoms evaluation details}
\label{app:kingdoms_eval}

\begin{figure}[!htb]
\centering
\includegraphics[width=0.55\linewidth]{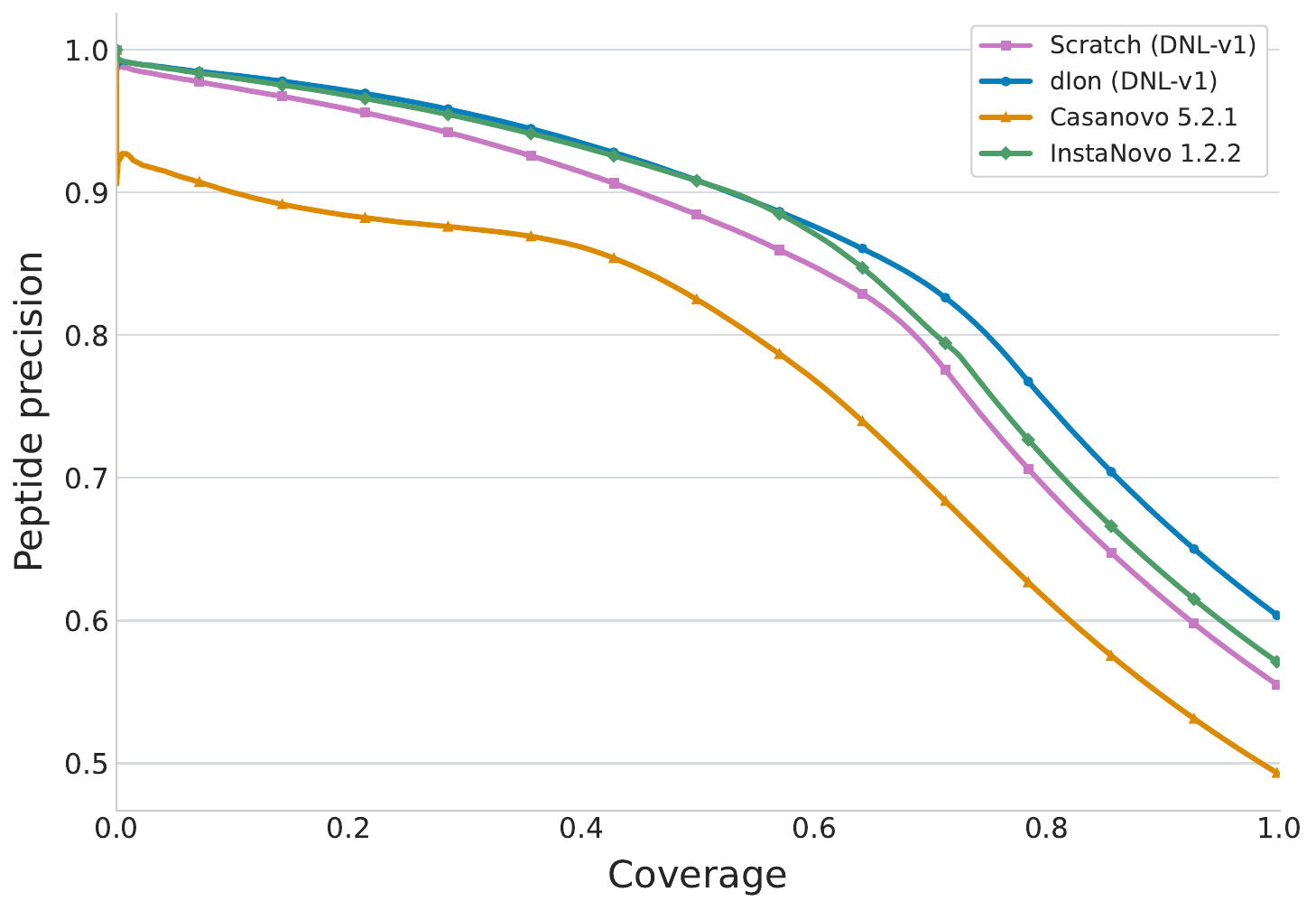}
\caption{Peptide precision against coverage on the
\DatasetShort{kingdomsdenovo} test set. Each model ranks its predictions by
its own confidence score, and coverage is the fraction of all test spectra
accepted, so the right endpoint is the full-coverage precision of
Table~\ref{tab:denovo-test}. \modelname{} and scratch are
fine-tuned end to end on \DatasetShort{kitchensinkvfive} with 200 peaks.}
\label{fig:denovo-precision-coverage}
\end{figure}

Table~\ref{tab:denovo-kingdoms-capped-complete} reports the complete Kingdoms
results, including amino-acid metrics and, for both initializations, models
fine-tuned with a 1,000-peak cap from the same 200-peak pretrained encoder.
With the larger cap, \modelname{} initialization gains in both training
regimes. Scratch training gains when trained on \DatasetShort{mskbfinal} and
not when trained on \DatasetShort{kitchensinkvfive}.

The canonical Kingdoms test set contains 4,926,232 unique spectra. For the
two \DatasetShort{kitchensinkvfive}-trained 200-peak models, peptide precision
is rescored from their complete prediction tables over exactly these spectra,
overall and per species, with the same matcher used for Casanovo and InstaNovo.
The other six in-house configurations, and the amino-acid metrics of all
eight, come from a distributed fast-evaluation mode. That mode padded the final batch
across sixteen workers by repeating the first eight records, all belonging to
\textit{Akkermansia muciniphila}: logged per-species counts sum to 4,926,240,
and that species shows 27,286 logged evaluations against 27,278 unique
spectra. The external Casanovo and InstaNovo results come from separate
sharded inference pipelines and do not share this behaviour. Nor does anything
on the \DatasetShort{mskbfinal} test set, where every model wrote individual
predictions to disk instead. The logged scores are reported unchanged. The effect is bounded and negligible: for a binary per-spectrum
metric such as peptide precision the difference between the logged and unique
denominators cannot exceed $8/4{,}926{,}240 \approx 1.6 \times 10^{-4}$
percentage points overall, or $8/27{,}286 \approx 0.03$ percentage points for
\textit{Akkermansia muciniphila} alone. No other species is affected.

Casanovo 5.2.1 supports precursor charges 1--4 and emitted predictions for
4,879,442 spectra. The 13,504 charge-$>4$ spectra and 33,286 supported spectra
without an emitted PSM are retained as incorrect, zero-confidence predictions,
so its full-coverage precision and precision--coverage curve use the complete
4,926,232-spectrum denominator. InstaNovo 1.2.2 emitted a prediction for every
spectrum.

\input{code/generated/tables/test_denovo_kingdoms_capped_complete}

Table~\ref{tab:denovo-kingdoms-per-species} breaks the full-coverage peptide
precision down by species for the two \DatasetShort{kitchensinkvfive}-trained
200-peak models, Casanovo 5.2.1 and InstaNovo 1.2.2. \modelname{}
initialization has the highest precision of the four models on every species,
by a margin that rounds to zero only on \textit{Rattus norvegicus}.

\input{code/generated/tables/test_denovo_kingdoms_per_species}

\subsection{\DatasetShort{mskbfinal} evaluation details}
\label{app:mskb_eval}

Table~\ref{tab:denovo-mskb-complete} reports the complete
\DatasetShort{mskbfinal} results, including amino-acid metrics, the models
fine-tuned with a 1,000-peak cap, and InstaNovo 1.2.2, which the main text
omits on this benchmark.

\input{code/generated/tables/test_denovo_mskb_complete}

\subsection{Oracle-precursor decoding of DIA spectra}
\label{app:dia_oracle}

In data-independent acquisition (DIA), wide isolation windows co-fragment
several peptides, so each MS2 spectrum is a mixture. As a pilot, we test
whether precursor conditioning can select one component of such a mixture.
Each query pairs a real wide-window DIA MS2 scan, with isolation windows at
least 26~$m/z$ wide and its 100 most intense peaks retained, with the precursor $m/z$ and charge of one peptide that
DIA-NN identified in it \citep{demichev2020diann}, and the model decodes that
peptide autoregressively. A scan recurs in several queries with different
precursors, giving 17,781 queries over 9,973 scans. The DIA-NN precursor acts
as an oracle selection query, so the pilot tests decoding of a supplied
component rather than MS1 precursor detection or full deconvolution.

The encoder is the Gram-refined \modelname{} encoder
(Appendix~\ref{app:denovo_refinement}), held frozen. The de novo decoder is
trained on synthetic two-spectrum mixtures of \DatasetShort{kitchensinkvfive},
built with the pretraining mixture construction: each mixture is decoded twice,
conditioned on either precursor, with that precursor's peptide as the target.
Neither pretraining nor decoder training uses DIA data. Inference uses
precursor conditioning and greedy decoding, and every query receives a
prediction. Table~\ref{tab:denovo-dia-oracle} reports the results. No baseline
has been evaluated on these queries. The absolute precision is low, but it is
exact recovery of target peptides from real DIA mixtures by a model trained
only on synthetic two-source mixtures of DDA spectra. A decisive test would
replace the oracle precursor with observed MS1 precursor features and use
narrower isolation windows.

\input{code/generated/tables/test_denovo_dia_oracle}

\subsection{Decoding and sequence evaluation}
\label{app:denovo_decoding}

All reported models use one frozen decoding protocol, summarized in
Table~\ref{tab:denovo-decoding}. Peptides are decoded in reverse order with
single-beam (greedy) decoding under a precursor-mass constraint.

\begin{table}[!htb]
\centering
\small
\caption{De novo decoding protocol, identical across all reported models.}
\label{tab:denovo-decoding}
\begin{tabular}{@{}ll@{}}
\toprule
Setting & Value \\
\midrule
Decoding order & Reverse peptide order \\
Search & Single beam (greedy) \\
Maximum peptide length & 100 \\
Minimum peptide length & 6 \\
Precursor mass tolerance & 50~ppm \\
Isotope-error range & 0--1 \\
\bottomrule
\end{tabular}
\end{table}

\textbf{Sequence matching.} Every model, including Casanovo and InstaNovo,
is scored with one implementation: a vendored copy of the Casanovo evaluator
\citep{yilmaz2022casanovo}, which applies the matching criterion introduced by
DeepNovo \citep{tran2017deepnovo}. Predicted and reference peptides are first
mapped to one token vocabulary and one table of residue masses, with
isoleucine and leucine treated as identical, so that each model's modification
notation resolves to the same tokens. Residues are aligned from the N terminus
and, if the peptides do not fully match, again from the C terminus. A
predicted residue matches when its mass differs from the aligned reference
residue by less than 0.1~Da and the cumulative mass up to and including it
differs by less than 0.5~Da.

Amino-acid precision and recall divide the number of matched residues by the
number of predicted and reference residues, respectively. A peptide is correct
only when every aligned position matches, so a missing or extra residue makes
it incorrect. Peptide precision is the number of correct peptides divided by
the number of test spectra. Spectra without a prediction, including those an
external model does not support, count as incorrect. On
\DatasetShort{mskbfinal}, Casanovo 5.2.1 with its stock preprocessing receives
196,979 charge-1--4 spectra. Of these, 20 fall below its 20-peak minimum after
filtering and one yields no PSM, and all 21 count as incorrect.

\textbf{Peptide confidence.} Following Casanovo \citep{yilmaz2024casanovo},
the confidence of a predicted peptide is the mean softmax probability of its
emitted tokens, including the stop token. When decoding ends without a stop
token, a probability of zero takes its place. If the peptide does not match
the precursor mass within 50~ppm, allowing isotope errors of 0--1, the
confidence is reduced by 1, so every precursor-compatible prediction ranks
above every incompatible one. Casanovo and InstaNovo keep their native
confidence scores. Scores only rank the predictions of one model and are not
comparable across models.

A precision--coverage curve sorts one model's predictions by decreasing
confidence. At rank $k$, coverage is $k$ divided by the number of test spectra
and precision is the fraction of the top $k$ predictions that are correct.
Spectra without a prediction rank last as errors, so the curve ends at the
full-coverage peptide precision.

\subsection{Online mini-de-novo probe}
\label{app:denovo_probe}

The online mini-de-novo probe tracks how readily the dense peak-token
representations support sequence prediction during pretraining. At each
evaluation the EMA-teacher encoder is frozen and its per-peak outputs are
cached, and a fresh two-layer Transformer decoder (two attention heads,
feed-forward width 128) is trained on them for four epochs with AdamW
(learning rate $3\times10^{-4}$, batch size 32). The probe data are drawn from
the \DatasetShort{mskbfinal} training set, restricted to unmodified peptides of
6 to 10 residues over 16 amino acids, and split by run and peptide into 7,000
training, 900 validation and 900 test spectra. Decoding and scoring follow
Appendix~\ref{app:denovo_decoding}, and we report validation peptide precision.
The probe is a development diagnostic, separate from the de novo evaluation of
Section~\ref{sec:denovo}.

\begin{figure}[!htb]
\centering
\begin{subfigure}[t]{0.40\linewidth}
  \centering
  \includegraphics[width=\linewidth]{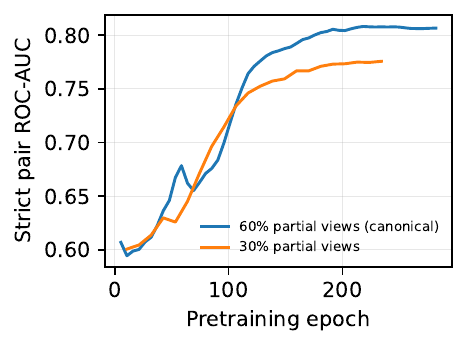}
  \caption{Strict pair ROC-AUC}
\end{subfigure}\hspace{0.03\linewidth}
\begin{subfigure}[t]{0.40\linewidth}
  \centering
  \includegraphics[width=\linewidth]{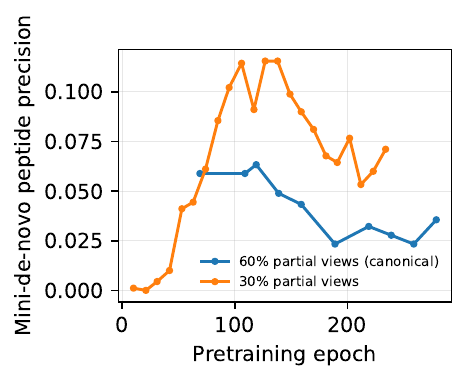}
  \caption{Mini-de-novo peptide precision}
\end{subfigure}
\caption{Partial views retaining 30\% or 60\% of peaks, \emph{validation} probes. The canonical run's mini-de-novo points are an offline reconstruction from its saved checkpoints.}
\label{fig:appendix-partial-view}
\end{figure}

Partial views retaining 30\% instead of 60\% of peaks give a much stronger
mini-de-novo probe, which peaks mid-training and then declines, but lower
strict pair ROC-AUC from mid-training onward
(Figure~\ref{fig:appendix-partial-view}). The two runs differ only in this
fraction, and the canonical recipe keeps 60\%. Its weaker frozen dense probe
does not carry over to fine-tuning: anchoring the canonical encoder to the 30\%
run's token features improves frozen de novo transfer but not end-to-end
fine-tuning, where the canonical encoder is already as good
(Appendix~\ref{app:denovo_refinement}).

\subsection{Gram-anchored refinement}
\label{app:denovo_refinement}

The mini-de-novo probe peaks mid-training and then declines, more steeply with
30\% partial views (Appendix~\ref{app:denovo_probe}). To recover the dense
information, we continued the canonical run from epoch 279 with Gram
anchoring \citep{simeoni2025dinov3}. Gram anchoring constrains the similarity
structure of the dense features rather than the features themselves. For a
spectrum with $N$ peaks, let $F \in \mathbb{R}^{N \times d}$ hold the
$\ell_2$-normalized peak-token outputs of the student and $F_a$ those of a
frozen anchor encoder. The Gram matrices $FF^\top$ and $F_aF_a^\top$ hold the
cosine similarity of every pair of peak tokens, and the loss is their mean
squared difference, $\lVert FF^\top - F_aF_a^\top \rVert_F^2 / N^2$. Because a
Gram matrix is unchanged by rotations of the feature space, the student remains
free to move its features under the DINO objective as long as it keeps the
anchor's pattern of which peaks resemble which. The anchor is the teacher of
the 30\% run at epoch 109, near the peak of its probe. The loss
is added to the DINO objective at weight 2.0, ramped in over 1,000 steps, with a
fixed learning rate of $4.5\times10^{-5}$ and a fixed EMA momentum of 0.9986.
Refinement ran for 45 epochs, and the epoch-298 checkpoint, where the probe
peaked, is used downstream.

\begin{figure}[!htb]
\centering
\begin{subfigure}[t]{0.40\linewidth}
  \centering
  \includegraphics[width=\linewidth]{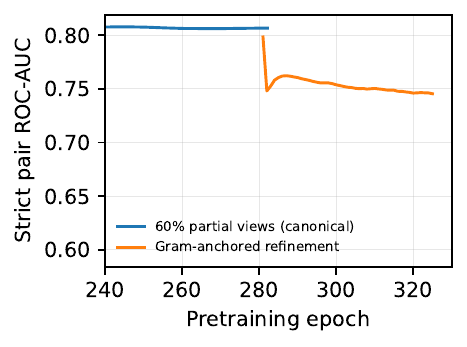}
  \caption{Strict pair ROC-AUC}
\end{subfigure}\hspace{0.03\linewidth}
\begin{subfigure}[t]{0.40\linewidth}
  \centering
  \includegraphics[width=\linewidth]{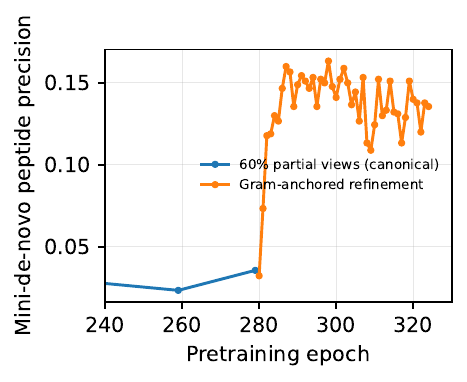}
  \caption{Mini-de-novo peptide precision}
\end{subfigure}
\caption{Gram-anchored refinement continued from epoch 279 of the canonical run, \emph{validation} probes. The canonical run's mini-de-novo points are an offline reconstruction from its saved checkpoints.}
\label{fig:appendix-gram}
\end{figure}

Refinement raises the mini-de-novo probe well above the unrefined run but
lowers strict pair ROC-AUC (Figure~\ref{fig:appendix-gram}). In de novo
transfer it substantially improves the frozen encoder, but end-to-end
fine-tuning of the canonical encoder is already as good
(Table~\ref{tab:gram-refinement}), so the main results use the unrefined
checkpoint. The refined encoder was fine-tuned end to end only on
\DatasetShort{kitchensinkvfive}, which leaves the \DatasetShort{mskbfinal}
end-to-end entry of Table~\ref{tab:gram-refinement} empty. The frozen DIA pilot of Appendix~\ref{app:dia_oracle} uses the
Gram-refined epoch-298 encoder.

\input{code/generated/tables/test_denovo_gram_refinement}

%% file: code/generated/tables/test_denovo_yeast_unlabeled_complete.tex
\begin{table}[!htb]
\centering
\small
\setlength{\tabcolsep}{3pt}
\caption{Complete unlabeled yeast results. Coverage is the fraction of submitted spectra receiving a prediction, tryptic restricts matches to tryptic peptides, the shuffled decoy preserves each protein's amino-acid composition, and modified calls are predictions carrying at least one modification.}
\label{tab:denovo-yeast-unlabeled-complete}
\begin{threeparttable}
\begin{tabular}{@{}lS[table-format=1.3]S[table-format=1.3]S[table-format=1.3]S[table-format=1.3]S[table-format=1.3]S[table-format=1.3]S[table-format=1.3]@{}}
\toprule
 & \multicolumn{1}{c}{} & \multicolumn{3}{c}{Matched} & \multicolumn{2}{c}{Decoy} & \multicolumn{1}{c}{} \\
\cmidrule(lr){3-5}
\cmidrule(lr){6-7}
{Model} & {Coverage} & {Emitted} & {All spectra} & {Tryptic} & {Reversed} & {Shuffled} & {Modified calls} \\
\midrule
\multicolumn{8}{@{}l}{\itshape This work} \\
\modelname{} init., \DatasetShort{kitchensinkvfive} training & \textbf{1.000} & \textbf{0.332} & \textbf{0.332} & \textbf{0.319} & 0.0025 & 0.0021 & 0.239 \\
\addlinespace
\multicolumn{8}{@{}l}{\itshape External reference} \\
Casanovo 5.2.1 & 0.900 & 0.324 & 0.292 & 0.311 & 0.0020 & 0.0024 & \textbf{0.326} \\
InstaNovo 1.2.2 & \textbf{1.000} & 0.320 & 0.320 & 0.305 & 0.0020 & 0.0017 & 0.291 \\
\bottomrule
\end{tabular}
\end{threeparttable}
\end{table}

%% file: code/generated/tables/test_denovo_kingdoms_capped_complete.tex
\begin{table}[!htb]
\centering
\small
\caption{Complete Kingdoms de novo results, including amino-acid metrics and the models fine-tuned with a 1,000-peak cap for both initializations. Within each training regime, scratch and \modelname{} initialization use an identical end-to-end training protocol.}
\label{tab:denovo-kingdoms-capped-complete}
\begin{threeparttable}
\begin{tabular}{@{}lS[table-format=1.3]S[table-format=1.3]S[table-format=1.3]@{}}
\toprule
 & \multicolumn{2}{c}{Amino acids} & \multicolumn{1}{c}{} \\
\cmidrule(lr){2-3}
{Training / initialization} & {Precision} & {Recall} & {Precision @ 100\% coverage} \\
\midrule
\multicolumn{4}{@{}l}{\itshape MSKB-final training} \\
Scratch, end-to-end & 0.610 & 0.612 & 0.449 \\
Scratch, end-to-end (1,000 peaks) & 0.615 & 0.617 & 0.454 \\
\modelname{} init., end-to-end & 0.702 & 0.702 & 0.533 \\
\modelname{} init., end-to-end (1,000 peaks) & 0.709 & 0.708 & 0.538 \\
\addlinespace
\multicolumn{4}{@{}l}{\itshape \DatasetShort{kitchensinkvfive} training} \\
Scratch, end-to-end & 0.703 & 0.704 & 0.554 \\
Scratch, end-to-end (1,000 peaks) & 0.701 & 0.704 & 0.553 \\
\modelname{} init., end-to-end & 0.755 & 0.755 & 0.602 \\
\modelname{} init., end-to-end (1,000 peaks) & \textbf{0.757} & \textbf{0.757} & \textbf{0.607} \\
\addlinespace
\multicolumn{4}{@{}l}{\itshape External reference} \\
Casanovo 5.2.1 & \multicolumn{1}{c}{\textemdash} & \multicolumn{1}{c}{\textemdash} & 0.492 \\
InstaNovo 1.2.2 & 0.696 & 0.698 & 0.570 \\
\bottomrule
\end{tabular}
\end{threeparttable}
\end{table}

%% file: code/generated/tables/test_denovo_kingdoms_per_species.tex
\begin{table}[p]
\centering
\scriptsize
\renewcommand{\arraystretch}{0.92}
\caption{Per-species peptide precision at full coverage on the \DatasetShort{kingdomsdenovo} test set (deterministic run-disjoint split, at most 100k spectra per species).}
\label{tab:denovo-kingdoms-per-species}
\begin{threeparttable}
\begin{tabular}{@{}lr*{4}{S[table-format=1.3]}@{}}
\toprule
 & & \multicolumn{2}{c}{\DatasetShort{kitchensinkvfive}, 200 peaks} & \multicolumn{2}{c}{External reference} \\
\cmidrule(lr){3-4}\cmidrule(lr){5-6}
{Species} & {Spectra} & {\modelname{} init.} & {Scratch} & {Casanovo 5.2.1} & {InstaNovo 1.2.2} \\
\midrule
\textit{Akkermansia muciniphila} & 27,278 & \textbf{0.584} & 0.535 & 0.487 & 0.546 \\
\textit{Arabidopsis thaliana} (callus) & 100,000 & \textbf{0.617} & 0.573 & 0.528 & 0.586 \\
\textit{Arabidopsis thaliana} (root) & 100,000 & \textbf{0.620} & 0.574 & 0.510 & 0.581 \\
\textit{Arabidopsis thaliana} (sprout) & 100,000 & \textbf{0.652} & 0.605 & 0.537 & 0.620 \\
\textit{Bacillus subtilis} & 17,126 & \textbf{0.559} & 0.527 & 0.506 & 0.546 \\
\textit{Bacteroides fragilis} & 76,935 & \textbf{0.631} & 0.583 & 0.505 & 0.574 \\
\textit{Bacteroides thetaiotaomicron} & 75,950 & \textbf{0.607} & 0.561 & 0.493 & 0.554 \\
\textit{Bacteroides uniformis} & 77,854 & \textbf{0.632} & 0.585 & 0.512 & 0.576 \\
\textit{Bacteroides vulgatus} & 69,903 & \textbf{0.612} & 0.562 & 0.491 & 0.554 \\
\textit{Bifidobacterium adolescentis} & 50,338 & \textbf{0.563} & 0.519 & 0.471 & 0.537 \\
\textit{Bifidobacterium longum} subsp.\ \textit{longum} & 40,502 & \textbf{0.525} & 0.487 & 0.451 & 0.503 \\
\textit{Blautia obeum} & 46,515 & \textbf{0.606} & 0.554 & 0.502 & 0.572 \\
\textit{Bos taurus} & 100,000 & \textbf{0.672} & 0.617 & 0.540 & 0.649 \\
\textit{Caenorhabditis elegans} & 100,000 & \textbf{0.633} & 0.584 & 0.519 & 0.600 \\
\textit{Canis lupus} & 100,000 & \textbf{0.631} & 0.576 & 0.504 & 0.607 \\
\textit{Chlamydomonas reinhardtii} & 45,545 & \textbf{0.537} & 0.504 & 0.476 & 0.520 \\
\textit{Chromera velia} & 100,000 & \textbf{0.535} & 0.492 & 0.450 & 0.519 \\
\textit{Clostridium bolteae} & 56,691 & \textbf{0.587} & 0.534 & 0.485 & 0.553 \\
\textit{Clostridium perfringens} & 59,588 & \textbf{0.642} & 0.594 & 0.538 & 0.600 \\
\textit{Clostridium saccharolyticum} & 61,717 & \textbf{0.608} & 0.556 & 0.502 & 0.570 \\
\textit{Collinsella aerofaciens} & 45,181 & \textbf{0.570} & 0.525 & 0.475 & 0.538 \\
\textit{Coprococcus comes} & 55,776 & \textbf{0.612} & 0.560 & 0.500 & 0.569 \\
\textit{Cricetulus griseus} & 100,000 & \textbf{0.551} & 0.500 & 0.451 & 0.545 \\
\textit{Danio rerio} & 100,000 & \textbf{0.603} & 0.554 & 0.485 & 0.575 \\
\textit{Dictyostelium discoideum} & 100,000 & \textbf{0.622} & 0.566 & 0.476 & 0.569 \\
\textit{Didelphis didelphinae} & 100,000 & \textbf{0.645} & 0.589 & 0.508 & 0.611 \\
\textit{Dorea formicigenerans} & 40,528 & \textbf{0.555} & 0.514 & 0.485 & 0.530 \\
\textit{Drosophila melanogaster} & 100,000 & \textbf{0.640} & 0.586 & 0.501 & 0.596 \\
\textit{Eggerthella lenta} & 25,395 & \textbf{0.519} & 0.462 & 0.408 & 0.482 \\
\textit{Emiliania huxleyi} & 100,000 & \textbf{0.560} & 0.515 & 0.449 & 0.530 \\
\textit{Escherichia coli} & 61,880 & \textbf{0.619} & 0.567 & 0.495 & 0.573 \\
\textit{Eubacterium rectale} & 54,888 & \textbf{0.564} & 0.516 & 0.467 & 0.521 \\
\textit{Fusarium oxysporum} & 100,000 & \textbf{0.494} & 0.457 & 0.438 & 0.476 \\
\textit{Fusobacterium nucleatum} subsp.\ \textit{nucleatum} & 60,505 & \textbf{0.641} & 0.594 & 0.538 & 0.599 \\
\textit{Gallus gallus} & 100,000 & \textbf{0.659} & 0.607 & 0.528 & 0.622 \\
\textit{Glycine max} & 100,000 & \textbf{0.608} & 0.561 & 0.494 & 0.573 \\
\textit{Gossypium hirsutum} & 100,000 & \textbf{0.607} & 0.560 & 0.504 & 0.575 \\
\textit{Guillardia theta} & 100,000 & \textbf{0.579} & 0.533 & 0.472 & 0.550 \\
\textit{Haloarcula marismortui} & 1,967 & \textbf{0.652} & 0.616 & 0.581 & 0.589 \\
\textit{Halobacterium} sp. & 10,815 & \textbf{0.556} & 0.522 & 0.489 & 0.518 \\
\textit{Homo sapiens} & 100,000 & \textbf{0.679} & 0.624 & 0.543 & 0.662 \\
\textit{Lactobacillus paracasei} & 51,446 & \textbf{0.586} & 0.541 & 0.489 & 0.554 \\
\textit{Mus musculus} & 100,000 & \textbf{0.611} & 0.555 & 0.479 & 0.596 \\
\textit{Mycoplasma intracellular} & 5,325 & \textbf{0.761} & 0.698 & 0.636 & 0.743 \\
\textit{Neurospora crassa} & 100,000 & \textbf{0.553} & 0.511 & 0.461 & 0.522 \\
\textit{Odoribacter splanchnicus} & 71,173 & \textbf{0.598} & 0.552 & 0.500 & 0.551 \\
\textit{Oryctolagus cuniculus} & 100,000 & \textbf{0.648} & 0.582 & 0.497 & 0.611 \\
\textit{Oryzias melastigma} & 100,000 & \textbf{0.593} & 0.539 & 0.471 & 0.557 \\
\textit{Parabacteroides distasonis} & 70,978 & \textbf{0.637} & 0.585 & 0.523 & 0.588 \\
\textit{Parabacteroides merdae} & 68,437 & \textbf{0.627} & 0.577 & 0.520 & 0.577 \\
\textit{Phaeodactylum tricornutum} & 100,000 & \textbf{0.555} & 0.511 & 0.457 & 0.532 \\
\textit{Porphyra umbilicalis} & 100,000 & \textbf{0.617} & 0.574 & 0.497 & 0.583 \\
\textit{Prevotella copri} & 10,287 & \textbf{0.646} & 0.598 & 0.537 & 0.599 \\
\textit{Rattus norvegicus} & 100,000 & \textbf{0.580} & 0.531 & 0.482 & \textbf{0.580} \\
\textit{Roseburia intestinalis} & 34,833 & \textbf{0.522} & 0.476 & 0.445 & 0.497 \\
\textit{Ruminococcus bromii} & 50,037 & \textbf{0.595} & 0.549 & 0.505 & 0.562 \\
\textit{Ruminococcus gnavus} & 49,639 & \textbf{0.623} & 0.573 & 0.505 & 0.578 \\
\textit{Ruminococcus torques} & 14,859 & \textbf{0.649} & 0.600 & 0.530 & 0.604 \\
\textit{Saccharomyces cerevisiae} & 100,000 & \textbf{0.650} & 0.602 & 0.517 & 0.602 \\
\textit{Stentor coeruleus} & 100,000 & \textbf{0.615} & 0.562 & 0.486 & 0.572 \\
\textit{Streptococcus parasanguinis} & 50,003 & \textbf{0.627} & 0.577 & 0.518 & 0.583 \\
\textit{Sulfolobus solfataricus} & 6,916 & \textbf{0.567} & 0.514 & 0.473 & 0.543 \\
\textit{Sus scrofa} & 100,000 & \textbf{0.577} & 0.525 & 0.462 & 0.572 \\
\textit{Symbiodinium microadriaticum} & 100,000 & \textbf{0.551} & 0.510 & 0.477 & 0.546 \\
\textit{Synechocystis nigrescens} & 2,324 & \textbf{0.652} & 0.615 & 0.577 & 0.633 \\
Tardigrade & 25,726 & \textbf{0.618} & 0.564 & 0.493 & 0.578 \\
\textit{Thalassiosira pseudonana} & 100,000 & \textbf{0.562} & 0.515 & 0.454 & 0.531 \\
\textit{Triticum aestivum} & 100,000 & \textbf{0.616} & 0.569 & 0.488 & 0.577 \\
\textit{Veillonella parvula} & 51,372 & \textbf{0.554} & 0.509 & 0.454 & 0.511 \\
\textit{Vitis vinifera} & 100,000 & \textbf{0.581} & 0.543 & 0.510 & 0.559 \\
\bottomrule
\end{tabular}
\end{threeparttable}
\end{table}

%% file: code/generated/tables/test_denovo_mskb_complete.tex
\begin{table}[!htb]
\centering
\small
\caption{Complete \DatasetShort{mskbfinal} de novo results, including amino-acid metrics and the \modelname{}-initialized models fine-tuned with a 1,000-peak cap. Within each training regime, scratch and \modelname{} initialization use an identical end-to-end training protocol.}
\label{tab:denovo-mskb-complete}
\begin{threeparttable}
\begin{tabular}{@{}lS[table-format=1.3]S[table-format=1.3]S[table-format=1.3]@{}}
\toprule
 & \multicolumn{2}{c}{Amino acids} & \multicolumn{1}{c}{} \\
\cmidrule(lr){2-3}
{Training / initialization} & {Precision} & {Recall} & {Precision @ 100\% coverage} \\
\midrule
\multicolumn{4}{@{}l}{\itshape MSKB-final training} \\
Scratch, end-to-end & 0.895 & 0.895 & 0.747 \\
\modelname{} init., end-to-end & 0.927 & 0.926 & 0.802 \\
\modelname{} init., end-to-end (1,000 peaks) & 0.935 & 0.934 & 0.822 \\
\addlinespace
\multicolumn{4}{@{}l}{\itshape \DatasetShort{kitchensinkvfive} training} \\
Scratch, end-to-end & 0.933 & 0.934 & 0.819 \\
\modelname{} init., end-to-end & 0.945 & 0.945 & 0.842 \\
\modelname{} init., end-to-end (1,000 peaks) & \textbf{0.949} & \textbf{0.949} & \textbf{0.856} \\
\addlinespace
\multicolumn{4}{@{}l}{\itshape External reference} \\
Casanovo 5.2.1 & 0.911 & 0.912 & 0.783 \\
\textcolor{black!55}{InstaNovo 1.2.2$^{*}$} & {\color{black!55}0.961} & {\color{black!55}0.961} & {\color{black!55}0.916} \\
\bottomrule
\end{tabular}
\begin{tablenotes}[flushleft]
\footnotesize
\item[] \textit{InstaNovo 1.2.2$^{*}$ (grey) is documented as training on MassIVE-KB v1, the source corpus of this test set. Its exact training subset and split assignments are unpublished, so overlap with the held-out partition cannot be excluded.}
\end{tablenotes}
\end{threeparttable}
\end{table}

%% file: code/generated/tables/test_denovo_dia_oracle.tex
\begin{table}[!htb]
\centering
\small
\caption{Oracle-precursor decoding of real wide-window DIA MS2 scans: 17,781 queries, each pairing one of 9,973 scans with the precursor of a DIA-NN-identified peptide in it. Metrics use the matcher of Appendix~\ref{app:denovo_decoding}.}
\label{tab:denovo-dia-oracle}
\begin{threeparttable}
\begin{tabular}{@{}lS[table-format=1.3]S[table-format=1.3]S[table-format=1.3]S[table-format=1.3]@{}}
\toprule
 & \multicolumn{2}{c}{Amino acids} & \multicolumn{2}{c}{} \\
\cmidrule(lr){2-3}
{Model} & {Precision} & {Recall} & {Peptide precision} & {Precision--coverage AUC} \\
\midrule
\modelname{}, frozen encoder & 0.225 & 0.227 & 0.106 & 0.328 \\
\bottomrule
\end{tabular}
\end{threeparttable}
\end{table}

%% file: code/generated/tables/test_denovo_gram_refinement.tex
\begin{table}[!htb]
\centering
\small
\caption{De novo peptide precision on the \DatasetShort{mskbfinal} test set with and without Gram-anchored refinement of the pretrained encoder.}
\label{tab:gram-refinement}
\begin{threeparttable}
\begin{tabular}{@{}lS[table-format=1.3]S[table-format=1.3]@{}}
\toprule
 & \multicolumn{2}{c}{Peptide precision} \\
\cmidrule(lr){2-3}
{Training corpus / encoder} & {Frozen encoder} & {End-to-end} \\
\midrule
\multicolumn{3}{@{}l}{\itshape \DatasetShort{kitchensinkvfive} training} \\
\modelname{} & 0.555 & 0.842 \\
\modelname{}, Gram-refined & 0.712 & 0.842 \\
\addlinespace
\multicolumn{3}{@{}l}{\itshape MSKB-final training} \\
\modelname{} & 0.487 & 0.802 \\
\modelname{}, Gram-refined & 0.658 & \multicolumn{1}{c}{\textemdash} \\
\bottomrule
\end{tabular}
\end{threeparttable}
\end{table}

%% file: appendix/F_training_diagnostics.tex
\section{Additional training diagnostics}
\label{app:training_diagnostics}

\subsection{Collapse monitoring}
\label{app:collapse}

Pretraining logs three diagnostics from the collapse analysis of DINO
\citep{caron2021dino}: the mean entropy $H(q)$ of the teacher assignments in
bits, the Kullback--Leibler divergence $D_{\mathrm{KL}}(q \,\|\, p)$ between
teacher and student assignments, and the largest across-batch variance of any
student output probability. A collapsed model gives the same output for every
input, which drives the divergence and the variance to zero. The loss alone
does not reveal this.

\begin{figure}[!htb]
\centering
\begin{subfigure}[t]{0.40\linewidth}
  \centering
  \includegraphics[width=\linewidth]{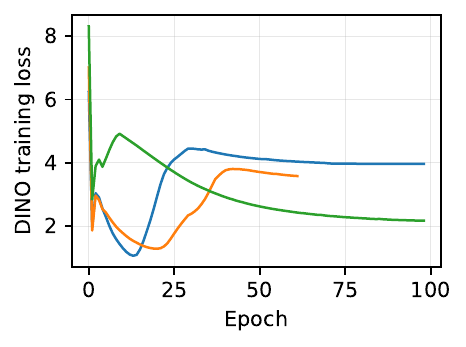}
  \caption{DINO training loss}
\end{subfigure}\hspace{0.03\linewidth}
\begin{subfigure}[t]{0.40\linewidth}
  \centering
  \includegraphics[width=\linewidth]{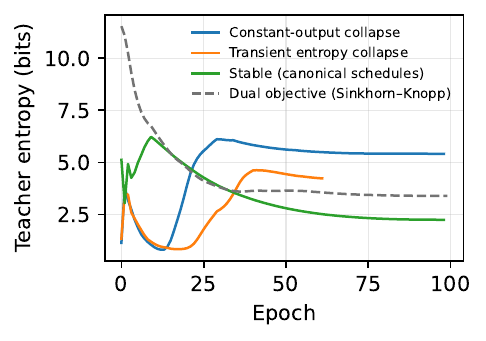}
  \caption{Teacher entropy}
\end{subfigure}

\vspace{0.5em}
\begin{subfigure}[t]{0.40\linewidth}
  \centering
  \includegraphics[width=\linewidth]{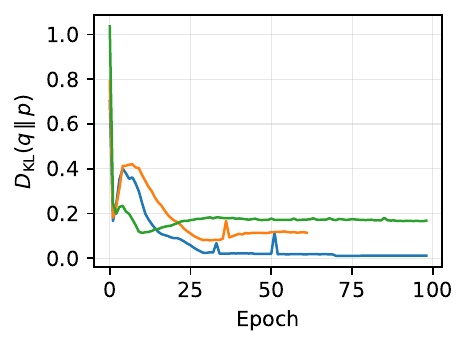}
  \caption{Teacher--student divergence}
\end{subfigure}\hspace{0.03\linewidth}
\begin{subfigure}[t]{0.40\linewidth}
  \centering
  \includegraphics[width=\linewidth]{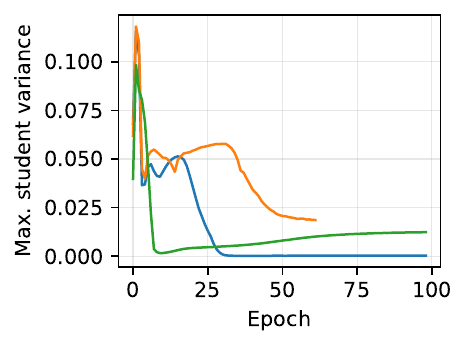}
  \caption{Maximum student variance}
\end{subfigure}
\caption{Collapse diagnostics of the three development runs and, in panel (b),
the 100-epoch dual-objective ablation run, as per-epoch means. For the
dual-objective run the entropy is that of the sharpened teacher distribution
before Sinkhorn--Knopp balancing. The transient-collapse run was stopped early.}
\label{fig:collapse}
\end{figure}

We chose the teacher temperature and momentum schedules on three early
development runs (Figure~\ref{fig:collapse}). These runs used the teacher
centering of the original DINO \citep{caron2021dino} instead of the
Sinkhorn--Knopp balancing of DINOv2 \citep{oquab2024dinov2}, and a naive early
objective with random peak subsets as views instead of the mixture and
partial-view tasks. The schedules
they selected nonetheless carried over unchanged to the dual objective. Their
two global views kept 95\% of the peaks and their five local views 20\%. Every run warms the teacher temperature up from
0.03. With a final temperature of 0.09, a 30-epoch warmup and a teacher
momentum rising from 0.996 to 0.9995, student and teacher converge to one
output distribution. The divergence and the variance fall to near zero while
$H(q)$ stays moderate, so the entropy alone does not detect this collapse. With
a final temperature of 0.07, the same slow warmup and a momentum rising from
0.997, the teacher instead over-sharpens, and $H(q)$ dips toward zero for
several epochs before partially recovering. A 10-epoch warmup to 0.07 with a
momentum rising from 0.998 to 1.0 keeps all three diagnostics away from their
degenerate values. These are the schedules of the canonical
recipe (Table~\ref{tab:ssl-optimization}), and the switch to Sinkhorn--Knopp
balancing, with the settings of DINOv2, required no further tuning. With these
schedules and Sinkhorn--Knopp balancing, the teacher entropy of the 100-epoch
dual-objective ablation run (Figure~\ref{fig:objective-ablation}) falls from
near-uniform at initialization to an intermediate value and stays well away
from both collapsed extremes (Figure~\ref{fig:collapse}, panel b).

\subsection{Masked-token objective screen}
\label{app:ibot_screen}

\begin{figure}[!htb]
\centering
\begin{subfigure}[t]{0.40\linewidth}
  \centering
  \includegraphics[width=\linewidth]{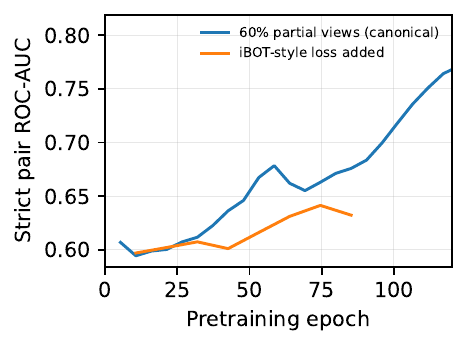}
  \caption{Strict pair ROC-AUC}
\end{subfigure}\hspace{0.03\linewidth}
\begin{subfigure}[t]{0.40\linewidth}
  \centering
  \includegraphics[width=\linewidth]{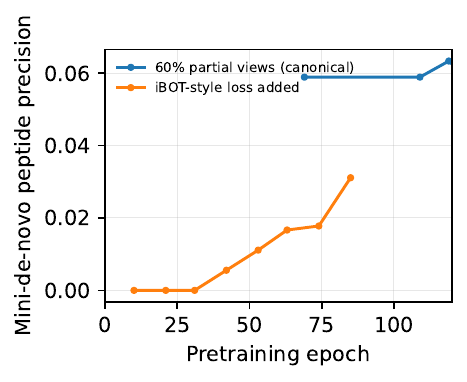}
  \caption{Mini-de-novo peptide precision}
\end{subfigure}
\caption{iBOT-style masked-token screen against the canonical run, \emph{validation} probes. The canonical run's mini-de-novo points are an offline reconstruction from its saved checkpoints.}
\label{fig:appendix-ibot}
\end{figure}

We screened adding an iBOT-style masked-token loss \citep{zhou2022ibot} to the
dual objective. The screen adds two precursor-conditioned global student views
in which 10--50\% of peak slots are replaced by a learned mask token, and
aligns the masked tokens with the teacher's clean token outputs at weight 0.5,
with all other settings as in the canonical recipe. The run stopped at epoch 85
of a nominal 300. Compared with the canonical run at matched epochs, it
reached lower strict pair ROC-AUC and lower mini-de-novo precision
(Appendix~\ref{app:denovo_probe}, Figure~\ref{fig:appendix-ibot}), so the loss
was not carried forward.

%% file: main.bbl
\begin{thebibliography}{56}
\providecommand{\natexlab}[1]{#1}
\providecommand{\url}[1]{\texttt{#1}}
\expandafter\ifx\csname urlstyle\endcsname\relax
  \providecommand{\doi}[1]{doi: #1}\else
  \providecommand{\doi}{doi: \begingroup \urlstyle{rm}\Url}\fi

\bibitem[Agrawal et~al.(2022)Agrawal, Mondal, Ghosh, and
  Richards]{agrawal2022alphareq}
Kumar~K. Agrawal, Arnab~Kumar Mondal, Arna Ghosh, and Blake Richards.
\newblock {$\alpha$-ReQ}: Assessing representation quality in self-supervised
  learning by measuring eigenspectrum decay.
\newblock In S.~Koyejo, S.~Mohamed, A.~Agarwal, D.~Belgrave, K.~Cho, and A.~Oh
  (eds.), \emph{Advances in Neural Information Processing Systems}, volume~35,
  pp.\  17626--17638. Curran Associates, Inc., 2022.
\newblock \doi{10.52202/068431-1281}.
\newblock URL
  \url{https://proceedings.neurips.cc/paper_files/paper/2022/hash/70596d70542c51c8d9b4e423f4bf2736-Abstract-Conference.html}.

\bibitem[Alex et~al.(2025)Alex, Atito, Mustafa, Awais, and
  Jackson]{alex2025sslam}
Tony Alex, Sara Atito, Armin Mustafa, Muhammad Awais, and Philip J.~B. Jackson.
\newblock {SSLAM}: Enhancing self-supervised models with audio mixtures for
  polyphonic soundscapes.
\newblock In \emph{International Conference on Learning Representations
  (ICLR)}, 2025.
\newblock URL \url{https://openreview.net/forum?id=odU59TxdiB}.

\bibitem[Altenburg et~al.(2026)Altenburg, Muth, van Zalm, Steen, and
  Renard]{altenburg2026yhydra}
Tom Altenburg, Thilo Muth, Patrick van Zalm, Hanno Steen, and Bernhard~Y.
  Renard.
\newblock Foundation model enables interpretable open and error-tolerant
  searching for mass spectrometry-based proteomics.
\newblock \emph{Bioinformatics}, 42\penalty0 (Supplement\_1):\penalty0 btag297,
  2026.
\newblock \doi{10.1093/bioinformatics/btag297}.

\bibitem[Assran et~al.(2022)Assran, Caron, Misra, Bojanowski, Bordes, Vincent,
  Joulin, Rabbat, and Ballas]{assran2022msn}
Mahmoud Assran, Mathilde Caron, Ishan Misra, Piotr Bojanowski, Florian Bordes,
  Pascal Vincent, Armand Joulin, Mike Rabbat, and Nicolas Ballas.
\newblock Masked {Siamese} networks for label-efficient learning.
\newblock In \emph{European Conference on Computer Vision (ECCV)}, pp.\
  456--473, 2022.
\newblock \doi{10.1007/978-3-031-19821-2_26}.

\bibitem[Assran et~al.(2023)Assran, Duval, Misra, Bojanowski, Vincent, Rabbat,
  LeCun, and Ballas]{assran2023ijepa}
Mahmoud Assran, Quentin Duval, Ishan Misra, Piotr Bojanowski, Pascal Vincent,
  Michael Rabbat, Yann LeCun, and Nicolas Ballas.
\newblock Self-supervised learning from images with a joint-embedding
  predictive architecture.
\newblock In \emph{Proceedings of the IEEE/CVF Conference on Computer Vision
  and Pattern Recognition (CVPR)}, pp.\  15619--15629, June 2023.

\bibitem[Baevski et~al.(2022)Baevski, Hsu, Xu, Babu, Gu, and
  Auli]{baevski2022data2vec}
Alexei Baevski, Wei-Ning Hsu, Qiantong Xu, Arun Babu, Jiatao Gu, and Michael
  Auli.
\newblock {data2vec}: A general framework for self-supervised learning in
  speech, vision and language.
\newblock In Kamalika Chaudhuri, Stefanie Jegelka, Le~Song, Csaba Szepesvari,
  Gang Niu, and Sivan Sabato (eds.), \emph{Proceedings of the 39th
  International Conference on Machine Learning}, volume 162 of
  \emph{Proceedings of Machine Learning Research}, pp.\  1298--1312. PMLR,
  17--23 Jul 2022.
\newblock URL \url{https://proceedings.mlr.press/v162/baevski22a.html}.

\bibitem[Bardes et~al.(2022)Bardes, Ponce, and LeCun]{bardes2022vicreg}
Adrien Bardes, Jean Ponce, and Yann LeCun.
\newblock {VICReg}: Variance-invariance-covariance regularization for
  self-supervised learning.
\newblock In \emph{International Conference on Learning Representations
  (ICLR)}, 2022.

\bibitem[Bittremieux(2025)]{bittremieux2025massivekb30m}
Wout Bittremieux.
\newblock {MassIVE-KB} v1 30 million {PSMs} training/validation/test splits,
  2025.

\bibitem[Bittremieux et~al.(2022)Bittremieux, May, Bilmes, and
  Noble]{bittremieux2022gleams}
Wout Bittremieux, Damon~H. May, Jeffrey Bilmes, and William~Stafford Noble.
\newblock A learned embedding for efficient joint analysis of millions of mass
  spectra.
\newblock \emph{Nature Methods}, 19:\penalty0 675--678, 2022.
\newblock \doi{10.1038/s41592-022-01496-1}.

\bibitem[Bushuiev et~al.(2026)Bushuiev, Bushuiev, Samusevich, Brungs, Sivic,
  and Pluskal]{bushuiev2026dreams}
Roman Bushuiev, Anton Bushuiev, Raman Samusevich, Corinna Brungs, Josef Sivic,
  and Tom{\'a}{\v{s}} Pluskal.
\newblock Self-supervised learning of molecular representations from millions
  of tandem mass spectra using {DreaMS}.
\newblock \emph{Nature Biotechnology}, 44\penalty0 (4):\penalty0 630--640,
  2026.
\newblock \doi{10.1038/s41587-025-02663-3}.

\bibitem[Caron et~al.(2020)Caron, Misra, Mairal, Goyal, Bojanowski, and
  Joulin]{caron2020swav}
Mathilde Caron, Ishan Misra, Julien Mairal, Priya Goyal, Piotr Bojanowski, and
  Armand Joulin.
\newblock Unsupervised learning of visual features by contrasting cluster
  assignments.
\newblock In \emph{Advances in Neural Information Processing Systems
  (NeurIPS)}, 2020.

\bibitem[Caron et~al.(2021)Caron, Touvron, Misra, J{\'e}gou, Mairal,
  Bojanowski, and Joulin]{caron2021dino}
Mathilde Caron, Hugo Touvron, Ishan Misra, Herv{\'e} J{\'e}gou, Julien Mairal,
  Piotr Bojanowski, and Armand Joulin.
\newblock Emerging properties in self-supervised vision transformers.
\newblock In \emph{Proceedings of the IEEE/CVF International Conference on
  Computer Vision (ICCV)}, pp.\  9650--9660, October 2021.

\bibitem[Chen et~al.(2020)Chen, Kornblith, Norouzi, and Hinton]{chen2020simclr}
Ting Chen, Simon Kornblith, Mohammad Norouzi, and Geoffrey Hinton.
\newblock A simple framework for contrastive learning of visual
  representations.
\newblock In \emph{Proceedings of the 37th International Conference on Machine
  Learning (ICML)}, volume 119 of \emph{Proceedings of Machine Learning
  Research}, pp.\  1597--1607, 2020.

\bibitem[Chen \& He(2021)Chen and He]{chen2021simsiam}
Xinlei Chen and Kaiming He.
\newblock Exploring simple {Siamese} representation learning.
\newblock In \emph{Proceedings of the IEEE/CVF Conference on Computer Vision
  and Pattern Recognition (CVPR)}, pp.\  15750--15758, June 2021.

\bibitem[Demichev et~al.(2020)Demichev, Messner, Vernardis, Lilley, and
  Ralser]{demichev2020diann}
Vadim Demichev, Christoph~B. Messner, Spyros~I. Vernardis, Kathryn~S. Lilley,
  and Markus Ralser.
\newblock {DIA-NN}: neural networks and interference correction enable deep
  proteome coverage in high throughput.
\newblock \emph{Nature Methods}, 17\penalty0 (1):\penalty0 41--44, 2020.
\newblock \doi{10.1038/s41592-019-0638-x}.

\bibitem[Devlin et~al.(2019)Devlin, Chang, Lee, and Toutanova]{devlin2019bert}
Jacob Devlin, Ming-Wei Chang, Kenton Lee, and Kristina Toutanova.
\newblock {BERT}: Pre-training of deep bidirectional transformers for language
  understanding.
\newblock In \emph{Proceedings of the 2019 Conference of the North American
  Chapter of the Association for Computational Linguistics: Human Language
  Technologies (NAACL-HLT)}, pp.\  4171--4186, 2019.

\bibitem[Eloff et~al.(2025)Eloff, Kalogeropoulos, Mabona, Morell, Catzel,
  Rivera-de Torre, Berg~Jespersen, Williams, van Beljouw, Skwark, Laustsen,
  Brouns, Ljungars, Schoof, Van~Goey, auf~dem Keller, Beguir, Lopez~Carranza,
  and Jenkins]{eloff2025instanovo}
Kevin Eloff, Konstantinos Kalogeropoulos, Amandla Mabona, Oliver Morell, Rachel
  Catzel, Esperanza Rivera-de Torre, Jakob Berg~Jespersen, Wesley Williams, Sam
  P.~B. van Beljouw, Marcin~J. Skwark, Andreas~Hougaard Laustsen, Stan J.~J.
  Brouns, Anne Ljungars, Erwin~M. Schoof, Jeroen Van~Goey, Ulrich auf~dem
  Keller, Karim Beguir, Nicolas Lopez~Carranza, and Timothy~P. Jenkins.
\newblock {InstaNovo} enables diffusion-powered de novo peptide sequencing in
  large-scale proteomics experiments.
\newblock \emph{Nature Machine Intelligence}, 7:\penalty0 565--579, 2025.
\newblock \doi{10.1038/s42256-025-01019-5}.

\bibitem[Garrido et~al.(2023)Garrido, Balestriero, Najman, and
  LeCun]{pmlr-v202-garrido23a}
Quentin Garrido, Randall Balestriero, Laurent Najman, and Yann LeCun.
\newblock {R}ank{M}e: Assessing the downstream performance of pretrained
  self-supervised representations by their rank.
\newblock In Andreas Krause, Emma Brunskill, Kyunghyun Cho, Barbara Engelhardt,
  Sivan Sabato, and Jonathan Scarlett (eds.), \emph{Proceedings of the 40th
  International Conference on Machine Learning}, volume 202 of
  \emph{Proceedings of Machine Learning Research}, pp.\  10929--10974. PMLR,
  23--29 Jul 2023.
\newblock URL \url{https://proceedings.mlr.press/v202/garrido23a.html}.

\bibitem[Gessulat et~al.(2019)Gessulat, Schmidt, Zolg, Samaras, Schnatbaum,
  Zerweck, Knaute, Rechenberger, Delanghe, Huhmer, Reimer, Ehrlich, Aiche,
  Kuster, and Wilhelm]{gessulat2019prosit}
Siegfried Gessulat, Tobias Schmidt, Daniel~Paul Zolg, Patroklos Samaras,
  Karsten Schnatbaum, Johannes Zerweck, Tobias Knaute, Julia Rechenberger,
  Bernard Delanghe, Andreas Huhmer, Ulf Reimer, Hans-Christian Ehrlich, Stephan
  Aiche, Bernhard Kuster, and Mathias Wilhelm.
\newblock {Prosit}: proteome-wide prediction of peptide tandem mass spectra by
  deep learning.
\newblock \emph{Nature Methods}, 16:\penalty0 509--518, 2019.
\newblock \doi{10.1038/s41592-019-0426-7}.

\bibitem[Gorshkov et~al.(2019)Gorshkov, Bubis, Solovyeva, Gorshkov, and
  Kjeldsen]{gorshkov2019corona}
Vladimir Gorshkov, Julia~A. Bubis, Elizaveta~M. Solovyeva, Mikhail~V. Gorshkov,
  and Frank Kjeldsen.
\newblock Protein corona formed on silver nanoparticles in blood plasma is
  highly selective and resistant to physicochemical changes of the solution.
\newblock \emph{Environmental Science: Nano}, 6\penalty0 (4):\penalty0
  1089--1098, 2019.
\newblock \doi{10.1039/C8EN01054D}.

\bibitem[Grill et~al.(2020)Grill, Strub, Altch{\'e}, Tallec, Richemond,
  Buchatskaya, Doersch, Pires, Guo, Azar, Piot, Kavukcuoglu, Munos, and
  Valko]{grill2020byol}
Jean-Bastien Grill, Florian Strub, Florent Altch{\'e}, Corentin Tallec, Pierre
  Richemond, Elena Buchatskaya, Carl Doersch, Bernardo~Avila Pires, Zhaohan
  Guo, Mohammad~Gheshlaghi Azar, Bilal Piot, Koray Kavukcuoglu, R{\'e}mi Munos,
  and Michal Valko.
\newblock Bootstrap your own latent: A new approach to self-supervised
  learning.
\newblock In \emph{Advances in Neural Information Processing Systems
  (NeurIPS)}, volume~33, pp.\  21271--21284, 2020.

\bibitem[He \& Ozay(2022)He and Ozay]{pmlr-v162-he22c}
Bobby He and Mete Ozay.
\newblock Exploring the gap between collapsed \& whitened features in
  self-supervised learning.
\newblock In Kamalika Chaudhuri, Stefanie Jegelka, Le~Song, Csaba Szepesvari,
  Gang Niu, and Sivan Sabato (eds.), \emph{Proceedings of the 39th
  International Conference on Machine Learning}, volume 162 of
  \emph{Proceedings of Machine Learning Research}, pp.\  8613--8634. PMLR,
  17--23 Jul 2022.
\newblock URL \url{https://proceedings.mlr.press/v162/he22c.html}.

\bibitem[He et~al.(2020)He, Fan, Wu, Xie, and Girshick]{he2020moco}
Kaiming He, Haoqi Fan, Yuxin Wu, Saining Xie, and Ross Girshick.
\newblock Momentum contrast for unsupervised visual representation learning.
\newblock In \emph{Proceedings of the IEEE/CVF Conference on Computer Vision
  and Pattern Recognition (CVPR)}, pp.\  9729--9738, 2020.

\bibitem[He et~al.(2022)He, Chen, Xie, Li, Doll{\'a}r, and Girshick]{he2022mae}
Kaiming He, Xinlei Chen, Saining Xie, Yanghao Li, Piotr Doll{\'a}r, and Ross
  Girshick.
\newblock Masked autoencoders are scalable vision learners.
\newblock In \emph{Proceedings of the IEEE/CVF Conference on Computer Vision
  and Pattern Recognition (CVPR)}, pp.\  16000--16009, 2022.

\bibitem[Houel et~al.(2010)Houel, Abernathy, Renganathan, Meyer-Arendt, Ahn,
  and Old]{houel2010chimera}
Stephane Houel, Robert Abernathy, Kutralanathan Renganathan, Karen
  Meyer-Arendt, Natalie~G. Ahn, and William~M. Old.
\newblock Quantifying the impact of chimera {MS/MS} spectra on peptide
  identification in large-scale proteomics studies.
\newblock \emph{Journal of Proteome Research}, 9\penalty0 (8):\penalty0
  4152--4160, 2010.
\newblock \doi{10.1021/pr1003856}.

\bibitem[Jin et~al.(2024)Jin, Xu, Zhang, Ling, Dong, Ouyang, Gao, Chang, and
  Sun]{jin2024contranovo}
Zhi Jin, Sheng Xu, Xiang Zhang, Tianze Ling, Nanqing Dong, Wanli Ouyang,
  Zhiqiang Gao, Cheng Chang, and Siqi Sun.
\newblock {ContraNovo}: A contrastive learning approach to enhance de novo
  peptide sequencing.
\newblock In \emph{Proceedings of the AAAI Conference on Artificial
  Intelligence}, volume~38, pp.\  144--152, 2024.
\newblock \doi{10.1609/aaai.v38i1.27765}.

\bibitem[Khosla et~al.(2020)Khosla, Teterwak, Wang, Sarna, Tian, Isola,
  Maschinot, Liu, and Krishnan]{khosla2020supcon}
Prannay Khosla, Piotr Teterwak, Chen Wang, Aaron Sarna, Yonglong Tian, Phillip
  Isola, Aaron Maschinot, Ce~Liu, and Dilip Krishnan.
\newblock Supervised contrastive learning.
\newblock In \emph{Advances in Neural Information Processing Systems
  (NeurIPS)}, volume~33, pp.\  18661--18673, 2020.

\bibitem[Kim et~al.(2020)Kim, Lee, Jung, Yeo, Seo, Lee, Lim, Chang, Song, Yang,
  Kim, Jung, Ahn, and Kim]{kim2020tent4}
Dongwan Kim, Young-suk Lee, Soo-Jin Jung, Jinah Yeo, Jenny~J. Seo, Young-Yoon
  Lee, Jaechul Lim, Hyeshik Chang, Jaewon Song, Jihye Yang, Jong-Seo Kim,
  Guhung Jung, Kwangseok Ahn, and V.~Narry Kim.
\newblock Viral hijacking of the {TENT4--ZCCHC14} complex protects viral {RNAs}
  via mixed tailing.
\newblock \emph{Nature Structural \& Molecular Biology}, 27\penalty0
  (6):\penalty0 581--588, 2020.
\newblock \doi{10.1038/s41594-020-0427-3}.

\bibitem[Lapin et~al.(2025)Lapin, Nilsson, Wilhelm, and
  K{\"a}ll]{lapin2025pairwise}
Joel Lapin, Alfred Nilsson, Mathias Wilhelm, and Lukas K{\"a}ll.
\newblock Pairwise attention: Leveraging mass differences to enhance de novo
  sequencing of mass spectra.
\newblock \emph{Journal of Proteome Research}, 24\penalty0 (7):\penalty0
  3722--3730, 2025.
\newblock \doi{10.1021/acs.jproteome.5c00063}.

\bibitem[LeCun(2022)]{lecun2022path}
Yann LeCun.
\newblock A path towards autonomous machine intelligence.
\newblock OpenReview, 2022.
\newblock URL \url{https://openreview.net/forum?id=BZ5a1r-kVsf}.
\newblock Version 0.9.2, released June 27, 2022.

\bibitem[Lee et~al.(2022)Lee, Mitchell, Burnet, Wu, Jenson, Merkley, Nakayasu,
  Nicora, Jansson, Burnum-Johnson, and Payne]{lee2022soilmetaproteomics}
Joon-Yong Lee, Hugh~D. Mitchell, Meagan~C. Burnet, Ruonan Wu, Sarah~C. Jenson,
  Eric~D. Merkley, Ernesto~S. Nakayasu, Carrie~D. Nicora, Janet~K. Jansson,
  Kristin~E. Burnum-Johnson, and Samuel~H. Payne.
\newblock Uncovering hidden members and functions of the soil microbiome using
  \textit{De Novo} metaproteomics.
\newblock \emph{Journal of Proteome Research}, 21\penalty0 (8):\penalty0
  2023--2035, 2022.
\newblock \doi{10.1021/acs.jproteome.2c00334}.

\bibitem[Lee et~al.(2021)Lee, Zhu, Sohn, Li, Shin, and Lee]{lee2021imix}
Kibok Lee, Yian Zhu, Kihyuk Sohn, Chun-Liang Li, Jinwoo Shin, and Honglak Lee.
\newblock {i-Mix}: A domain-agnostic strategy for contrastive representation
  learning.
\newblock In \emph{International Conference on Learning Representations
  (ICLR)}, 2021.
\newblock URL \url{https://openreview.net/forum?id=T6AxtOaWydQ}.

\bibitem[Li et~al.(2024)Li, Kang, Baek, Kim, Park, and Jung]{li2024fkbpl}
Cathena~Meiling Li, Jaemin Kang, Jongyeon Baek, Youbin Kim, Heemin Park, and
  Yong-Keun Jung.
\newblock Cytosolic {FKBPL} and {ER}-resident {CKAP4} co-regulates {ER}-phagy
  and protein secretion.
\newblock \emph{Nature Communications}, 15, 2024.
\newblock \doi{10.1038/s41467-024-52188-7}.

\bibitem[Lim et~al.(2025)Lim, Lee, Ju, Kim, Mun, Jeon, Lee, Lee, Ku, Kim, Bae,
  Kim, and Kim]{lim2025dsrna}
JinA Lim, Namseok Lee, Seonmin Ju, Jeesoo Kim, Subin Mun, Moonhyeon Jeon,
  Yong-ki Lee, Seok-Hoon Lee, Jayoung Ku, Sujin Kim, Sangsu Bae, Jong-Seo Kim,
  and Yoosik Kim.
\newblock Cellular {dsRNA} interactome captured by {K1} antibody reveals the
  regulatory map of exogenous {RNA} sensing.
\newblock \emph{Communications Biology}, 8, 2025.
\newblock \doi{10.1038/s42003-025-07807-4}.

\bibitem[Melendez(2024)]{melendez2024mskbfinaldata}
Carlo Melendez.
\newblock Data for ``accounting for digestion enzyme bias in {Casanovo}'',
  2024.

\bibitem[Melendez et~al.(2024)Melendez, Sanders, Yilmaz, Bittremieux, Fondrie,
  Oh, and Noble]{melendez2024enzymebias}
Carlo Melendez, Justin Sanders, Melih Yilmaz, Wout Bittremieux, William~E.
  Fondrie, Sewoong Oh, and William~Stafford Noble.
\newblock Accounting for digestion enzyme bias in {Casanovo}.
\newblock \emph{Journal of Proteome Research}, 23\penalty0 (10):\penalty0
  4761--4769, 2024.
\newblock \doi{10.1021/acs.jproteome.4c00422}.

\bibitem[Michalski et~al.(2011)Michalski, Cox, and
  Mann]{michalski2011detectable}
Annette Michalski, J{\"u}rgen Cox, and Matthias Mann.
\newblock More than 100,000 detectable peptide species elute in single shotgun
  proteomics runs but the majority is inaccessible to data-dependent
  {LC-MS/MS}.
\newblock \emph{Journal of Proteome Research}, 10\penalty0 (4):\penalty0
  1785--1793, 2011.
\newblock \doi{10.1021/pr101060v}.

\bibitem[M{\"u}ller et~al.(2020)M{\"u}ller, Geyer, Cola{\c{c}}o, Treit,
  Strauss, Oroshi, Doll, Virreira~Winter, Bader, K{\"o}hler, Theis, Santos, and
  Mann]{muller2020kingdoms}
Johannes~B. M{\"u}ller, Philipp~E. Geyer, Ana~R. Cola{\c{c}}o, Peter~V. Treit,
  Maximilian~T. Strauss, Mario Oroshi, Sophia Doll, Sebastian Virreira~Winter,
  Jakob~M. Bader, Niklas K{\"o}hler, Fabian Theis, Alberto Santos, and Matthias
  Mann.
\newblock The proteome landscape of the kingdoms of life.
\newblock \emph{Nature}, 582\penalty0 (7813):\penalty0 592--596, 2020.
\newblock \doi{10.1038/s41586-020-2402-x}.

\bibitem[Nieuwoudt et~al.(2026)Nieuwoudt, Reverenna, Patel, Catzel, Houngue,
  Daniel, Eloff, Santos, Lopez~Carranza, Jenkins, Van~Goey, and
  Kalogeropoulos]{nieuwoudt2026instanovofm}
Mechiel Nieuwoudt, Marco Reverenna, Divanisha Patel, Rachel Catzel, Isaac H.~J.
  Houngue, Jemma Daniel, Kevin Eloff, Alberto Santos, Nicolas Lopez~Carranza,
  Timothy~P. Jenkins, Jeroen Van~Goey, and Konstantinos Kalogeropoulos.
\newblock Learning from tandem mass spectra at scale with a self-supervised
  foundation model for proteomics.
\newblock \emph{bioRxiv}, 2026.
\newblock \doi{10.64898/2026.09.03.747733}.

\bibitem[Oquab et~al.(2024)Oquab, Darcet, Moutakanni, Vo, Szafraniec, Khalidov,
  Fernandez, Haziza, Massa, El-Nouby, Assran, Ballas, Galuba, Howes, Huang, Li,
  Misra, Rabbat, Sharma, Synnaeve, Xu, J{\'e}gou, Mairal, Labatut, Joulin, and
  Bojanowski]{oquab2024dinov2}
Maxime Oquab, Timoth{\'e}e Darcet, Th{\'e}o Moutakanni, Huy~V. Vo, Marc
  Szafraniec, Vasil Khalidov, Pierre Fernandez, Daniel Haziza, Francisco Massa,
  Alaaeldin El-Nouby, Mido Assran, Nicolas Ballas, Wojciech Galuba, Russell
  Howes, Po-Yao Huang, Shang-Wen Li, Ishan Misra, Michael Rabbat, Vasu Sharma,
  Gabriel Synnaeve, Hu~Xu, Herv{\'e} J{\'e}gou, Julien Mairal, Patrick Labatut,
  Armand Joulin, and Piotr Bojanowski.
\newblock {DINOv2}: Learning robust visual features without supervision.
\newblock \emph{Transactions on Machine Learning Research}, 2024.

\bibitem[Perez-Riverol et~al.(2025)Perez-Riverol, Bandla, Kundu,
  Kamatchinathan, Bai, Hewapathirana, John, Prakash, Walzer, Wang, and
  Vizca{\'\i}no]{perezriverol2025pride}
Yasset Perez-Riverol, Chakradhar Bandla, Deepti~J. Kundu, Selvakumar
  Kamatchinathan, Jingwen Bai, Suresh Hewapathirana, Nithu~Sara John, Ananth
  Prakash, Mathias Walzer, Shengbo Wang, and Juan~Antonio Vizca{\'\i}no.
\newblock The {PRIDE} database at 20 years: 2025 update.
\newblock \emph{Nucleic Acids Research}, 53\penalty0 (D1):\penalty0 D543--D553,
  2025.
\newblock \doi{10.1093/nar/gkae1011}.

\bibitem[Sanders et~al.(2025)Sanders, Yilmaz, Russell, Bittremieux, Fondrie,
  Riley, Oh, and Noble]{sanders2025foundation}
Justin Sanders, Melih Yilmaz, Jacob~H. Russell, Wout Bittremieux, William~E.
  Fondrie, Nicholas~M. Riley, Sewoong Oh, and William~Stafford Noble.
\newblock Foundation model for mass spectrometry proteomics.
\newblock \emph{arXiv preprint arXiv:2505.10848}, 2025.

\bibitem[Shen et~al.(2022)Shen, Liu, Liu, Savvides, Darrell, and
  Xing]{shen2022unmix}
Zhiqiang Shen, Zechun Liu, Zhuang Liu, Marios Savvides, Trevor Darrell, and
  Eric Xing.
\newblock {Un-Mix}: Rethinking image mixtures for unsupervised visual
  representation learning.
\newblock In \emph{Proceedings of the AAAI Conference on Artificial
  Intelligence}, volume~36, pp.\  2216--2224, 2022.
\newblock \doi{10.1609/aaai.v36i2.20119}.

\bibitem[Sim{\'e}oni et~al.(2025)Sim{\'e}oni, Vo, Seitzer, Baldassarre, Oquab,
  Jose, Khalidov, Szafraniec, Yi, Ramamonjisoa, Massa, Haziza, Wehrstedt, Wang,
  Darcet, Moutakanni, Sentana, Roberts, Vedaldi, Tolan, Brandt, Couprie,
  Mairal, J{\'e}gou, Labatut, and Bojanowski]{simeoni2025dinov3}
Oriane Sim{\'e}oni, Huy~V. Vo, Maximilian Seitzer, Federico Baldassarre, Maxime
  Oquab, Cijo Jose, Vasil Khalidov, Marc Szafraniec, Seungeun Yi, Micha{\"e}l
  Ramamonjisoa, Francisco Massa, Daniel Haziza, Luca Wehrstedt, Jianyuan Wang,
  Timoth{\'e}e Darcet, Th{\'e}o Moutakanni, Leonel Sentana, Claire Roberts,
  Andrea Vedaldi, Jamie Tolan, John Brandt, Camille Couprie, Julien Mairal,
  Herv{\'e} J{\'e}gou, Patrick Labatut, and Piotr Bojanowski.
\newblock {DINOv3}.
\newblock \emph{arXiv preprint arXiv:2508.10104}, 2025.

\bibitem[Tran et~al.(2017)Tran, Zhang, Xin, Shan, and Li]{tran2017deepnovo}
Ngoc~Hieu Tran, Xianglilan Zhang, Lei Xin, Baozhen Shan, and Ming Li.
\newblock De novo peptide sequencing by deep learning.
\newblock \emph{Proceedings of the National Academy of Sciences}, 114\penalty0
  (31):\penalty0 8247--8252, 2017.
\newblock \doi{10.1073/pnas.1705691114}.

\bibitem[Tsitsulin et~al.(2023)Tsitsulin, Munkhoeva, and
  Perozzi]{pmlr-v221-tsitsulin23a}
Anton Tsitsulin, Marina Munkhoeva, and Bryan Perozzi.
\newblock Unsupervised embedding quality evaluation.
\newblock In Timothy Doster, Tegan Emerson, Henry Kvinge, Nina Miolane,
  Mathilde Papillon, Bastian Rieck, and Sophia Sanborn (eds.),
  \emph{Proceedings of 2nd Annual Workshop on Topology, Algebra, and Geometry
  in Machine Learning (TAG-ML)}, volume 221 of \emph{Proceedings of Machine
  Learning Research}, pp.\  169--188. PMLR, 28 Jul 2023.
\newblock URL \url{https://proceedings.mlr.press/v221/tsitsulin23a.html}.

\bibitem[Vaswani et~al.(2017)Vaswani, Shazeer, Parmar, Uszkoreit, Jones, Gomez,
  Kaiser, and Polosukhin]{vaswani2017attention}
Ashish Vaswani, Noam Shazeer, Niki Parmar, Jakob Uszkoreit, Llion Jones,
  Aidan~N. Gomez, Lukasz Kaiser, and Illia Polosukhin.
\newblock Attention is all you need.
\newblock In \emph{Advances in Neural Information Processing Systems},
  volume~30, pp.\  5998--6008, 2017.

\bibitem[Wang et~al.(2018)Wang, Wang, Carver, Pullman, Cha, and
  Bandeira]{wang2018massivekb}
Mingxun Wang, Jian Wang, Jeremy Carver, Benjamin~S. Pullman, Seong~Won Cha, and
  Nuno Bandeira.
\newblock Assembling the community-scale discoverable human proteome.
\newblock \emph{Cell Systems}, 7\penalty0 (4):\penalty0 412--421.e5, 2018.
\newblock \doi{10.1016/j.cels.2018.08.004}.

\bibitem[Wang et~al.(2021)Wang, Zhang, Shen, Kong, and Li]{wang2021densecl}
Xinlong Wang, Rufeng Zhang, Chunhua Shen, Tao Kong, and Lei Li.
\newblock Dense contrastive learning for self-supervised visual pre-training.
\newblock In \emph{Proceedings of the IEEE/CVF Conference on Computer Vision
  and Pattern Recognition (CVPR)}, pp.\  3024--3033, 2021.

\bibitem[Yang et~al.(2024)Yang, Ling, Sun, Liang, Xu, Huang, Xie, He, Li, He,
  Wang, and Chang]{yang2024pihelixnovo}
Tingpeng Yang, Tianze Ling, Boyan Sun, Zhendong Liang, Fan Xu, Xiansong Huang,
  Linhai Xie, Yonghong He, Leyuan Li, Fuchu He, Yu~Wang, and Cheng Chang.
\newblock Introducing {$\pi$}-{HelixNovo} for practical large-scale de novo
  peptide sequencing.
\newblock \emph{Briefings in Bioinformatics}, 25\penalty0 (2):\penalty0
  bbae021, 2024.
\newblock \doi{10.1093/bib/bbae021}.

\bibitem[Yilmaz et~al.(2022)Yilmaz, Fondrie, Bittremieux, Oh, and
  Noble]{yilmaz2022casanovo}
Melih Yilmaz, William Fondrie, Wout Bittremieux, Sewoong Oh, and William~S
  Noble.
\newblock De novo mass spectrometry peptide sequencing with a transformer
  model.
\newblock In Kamalika Chaudhuri, Stefanie Jegelka, Le~Song, Csaba Szepesvari,
  Gang Niu, and Sivan Sabato (eds.), \emph{Proceedings of the 39th
  International Conference on Machine Learning}, volume 162 of
  \emph{Proceedings of Machine Learning Research}, pp.\  25514--25522. PMLR,
  17--23 Jul 2022.

\bibitem[Yilmaz et~al.(2024)Yilmaz, Fondrie, Bittremieux, Melendez, Nelson,
  Ananth, Oh, and Noble]{yilmaz2024casanovo}
Melih Yilmaz, William~E. Fondrie, Wout Bittremieux, Carlo~F. Melendez, Rowan
  Nelson, Varun Ananth, Sewoong Oh, and William~Stafford Noble.
\newblock Sequence-to-sequence translation from mass spectra to peptides with a
  transformer model.
\newblock \emph{Nature Communications}, 15:\penalty0 6427, 2024.
\newblock \doi{10.1038/s41467-024-49731-x}.

\bibitem[Zhang et~al.(2024)Zhang, Yang, Xie, Wang, Zhang, and
  Lu]{zhang2024msbert}
Hailiang Zhang, Qiong Yang, Ting Xie, Yue Wang, Zhimin Zhang, and Hongmei Lu.
\newblock {MSBERT}: Embedding tandem mass spectra into chemically rational
  space by mask learning and contrastive learning.
\newblock \emph{Analytical Chemistry}, 96\penalty0 (42):\penalty0 16599--16608,
  2024.
\newblock \doi{10.1021/acs.analchem.4c02426}.

\bibitem[Zhao et~al.(2026)Zhao, Mao, Wang, Li, Peng, Chen, Lu, Ji, Ding, Zhang,
  Liao, E, Wen, Zhang, and Chi]{zhao2026punifind}
Jiale Zhao, Peng-Zhi Mao, Kaifei Wang, Yi-Ming Li, Ya-Ping Peng, Ranfei Chen,
  Shuqi Lu, Xiaohong Ji, Jiaxiang Ding, Xin Zhang, Yucheng Liao, Weinan E, Han
  Wen, Wei-Jie Zhang, and Hao Chi.
\newblock A large-scale unified deep learning model for peptide mass spectrum
  interpretation trained on multimodal data.
\newblock \emph{Nature Machine Intelligence}, 8:\penalty0 997--1011, 2026.
\newblock \doi{10.1038/s42256-026-01234-8}.

\bibitem[Zhou et~al.(2022)Zhou, Wei, Wang, Shen, Xie, Yuille, and
  Kong]{zhou2022ibot}
Jinghao Zhou, Chen Wei, Huiyu Wang, Wei Shen, Cihang Xie, Alan Yuille, and Tao
  Kong.
\newblock {iBOT}: Image {BERT} pre-training with online tokenizer.
\newblock In \emph{International Conference on Learning Representations}, 2022.

\bibitem[Zolg et~al.(2017)Zolg, Wilhelm, Schnatbaum, Zerweck, Knaute, Delanghe,
  Bailey, Gessulat, Ehrlich, Weininger, Yu, Schlegl, Kramer, Schmidt,
  Kusebauch, Deutsch, Aebersold, Moritz, Wenschuh, Moehring, Aiche, Huhmer,
  Reimer, and Kuster]{zolg2017proteometools}
Daniel~P. Zolg, Mathias Wilhelm, Karsten Schnatbaum, Johannes Zerweck, Tobias
  Knaute, Bernard Delanghe, Derek~J. Bailey, Siegfried Gessulat, Hans-Christian
  Ehrlich, Maximilian Weininger, Peng Yu, Judith Schlegl, Karl Kramer, Tobias
  Schmidt, Ulrike Kusebauch, Eric~W. Deutsch, Ruedi Aebersold, Robert~L.
  Moritz, Holger Wenschuh, Thomas Moehring, Stephan Aiche, Andreas Huhmer, Ulf
  Reimer, and Bernhard Kuster.
\newblock Building {ProteomeTools} based on a complete synthetic human
  proteome.
\newblock \emph{Nature Methods}, 14\penalty0 (3):\penalty0 259--262, 2017.
\newblock \doi{10.1038/nmeth.4153}.

\end{thebibliography}
